\documentclass[preprint,12pt]{elsarticle}

\usepackage{amssymb}
\usepackage{algorithm}
\usepackage{algorithmic}
\usepackage{mathtools}
\usepackage{subcaption}
\usepackage{booktabs}
\usepackage{wrapfig}
\usepackage{setspace} 

\makeatletter
\def\ps@pprintTitle{%
  \let\@oddhead\@empty
  \let\@evenhead\@empty
  \let\@oddfoot\@empty
  \let\@evenfoot\@oddfoot
}
\makeatother
\begin{document}

\begin{frontmatter}

\title{Increasing-Margin Adversarial (IMA) Training to Improve Adversarial Robustness of Neural Networks}
\author[l3]{Linhai Ma}
\ead{l.ma@miami.edu}

 \author[l3]{Liang Liang\corref{cor1}}
 \ead{liang@cs.miami.edu}
 \cortext[cor1]{Corresponding Author}



\address[l3]{Department of Computer Science, University of Miami,
             1365 Memorial Drive,
             Coral Gables,
             33146,
             FL,
             USA}

\begin{abstract}
Background and Objective: Deep neural networks (DNNs) are vulnerable to adversarial noises. Adversarial training is a general and effective strategy to improve DNN robustness (i.e., accuracy on noisy data) against adversarial noises. However, DNN models trained by the current existing adversarial training methods may have much lower standard accuracy (i.e., accuracy on clean data), compared to the same models trained by the standard method on clean data, and this phenomenon is known as the trade-off between accuracy and robustness and is considered unavoidable. This issue prevents adversarial training from being used in many application domains, such as medical image analysis, as practitioners do not want to sacrifice standard accuracy too much in exchange for adversarial robustness. Our objective is to lift (i.e., alleviate or even avoid) this trade-off between standard accuracy and adversarial robustness for medical image classification and segmentation.
\par
Methods: We propose a novel adversarial training method, named Increasing-Margin Adversarial (IMA) Training, which is supported by an equilibrium state analysis about the optimality of adversarial training samples. Our method aims to preserve accuracy while improving robustness by generating optimal adversarial training samples. We evaluate our method and the other eight representative methods on six publicly available image datasets corrupted by noises generated by AutoAttack and white-noise attack.
\par
Results: Our method achieves the highest adversarial robustness for image classification and segmentation with the smallest reduction in accuracy on clean data. For one of the applications, our method improves both accuracy and robustness. The source code will be publicly available on GitHub when the paper is published.
\par
Conclusions: Our study has demonstrated that our method can lift the trade-off between standard accuracy and adversarial robustness for the image classification and segmentation applications. To our knowledge, it is the first work to show that the trade-off is avoidable for medical image segmentation.
\end{abstract}




\end{frontmatter}


\section{Introduction}
\label{1}

\subsection{Adversarial Robustness is Essential for Medical Applications}

\begin{figure}[ht]
\centering
        \centering
        \includegraphics[width=0.6\linewidth]{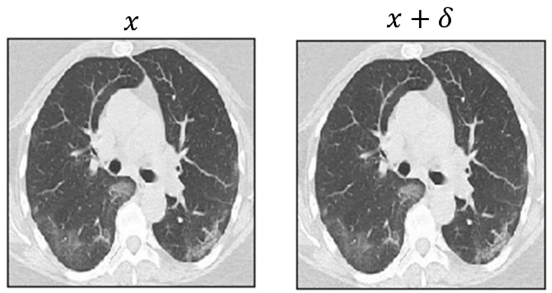}
        \centering
        \caption{An example of a clean image (left) and the image with unperceivable adversarial noise (right). We modify a ResNet-18 model and train it on a public COVID-19 CT image dataset (images are resized to 224$\times$224), and then test the model's robustness. $x$ denotes an image of a COVID-19-infected lung, which is correctly classified as infected (left). After adding an unperceivable adversarial noise $\delta$ to $x$, the noisy image $x+\delta$ is classified as uninfected (right). On the test set, although the model achieved $\ge 95\%$ accuracy on clean images, its accuracy dropped to zero on images with small adversarial noises (see Section 3 for more details).}
        \label{fig1}
\end{figure}

Deep neural networks (DNNs) have become the first choice for automated image analysis due to their superior performance. However, recent studies have shown that DNNs are very vulnerable to adversarial noises. Adversarial noises were first discovered by \cite{szegedy2014intriguing} and then explained by \cite{goodfellow2014explaining}. It is known that adversarial noises can significantly affect the robustness of DNNs for image-related applications \cite{akhtar2018threat, graese2016assessing, mirjalili2017soft, eykholt2018robust}. The COVID-19 pandemic has caused the death of millions of people \cite{WHO2020Coronavirus}. A large-scale study shows that CT had higher sensitivity for the diagnosis of COVID-19 as compared with initial reverse-transcription polymerase chain reaction (RT-PCR) from swab samples \cite{ai2020correlation}. As reviewed in \cite{shi2020review}, many DNN models for COVID-19 diagnosis from CT images have been developed and achieved very high classification accuracy. However, few of these studies \cite{shi2020review} considered DNN robustness against adversarial noises. Fig. \ref{fig1} shows that Resnet-18 model \cite{he2016deep} is very vulnerable to unperceivable adversarial noise, and this non-robust model cannot be trusted in clinical applications. 

It is well known that adversarial noises are the worst-case random noises \cite{fawzi2016robustness, gilmer2019adversarial}, and an adversarially-robust DNN model is also robust to random noises that may exist everywhere in the real world \cite{fawzi2016robustness}. This statement is also supported by our observation (see the Results Section). Therefore, adversarial noises/attacks are not just security issues caused by hackers, and the worst-case random imaging noises could also be ``adversarial" leading to wrong classifications. Thus, adversarial robustness should be a built-in property of every DNN for medical applications. The DNNs in the previous COVID-19 studies \cite{shi2020review} should be checked and enhanced for adversarial robustness before being deployed in clinics and hospitals. In short, the adversarial robustness problem is critical and should be thoroughly researched.

\subsection{Our Innovations and Contributions}

To improve the adversarial robustness of a DNN model, adversarial training is the most general strategy. By generating adversarial training samples to train the model, adversarial training can improve the adversarial robustness of the model. The standard adversarial training (SAT) \cite{madry2017towards, kurakin2016adversarial} generates adversarial training samples with a fixed and uniform adversarial noise upper bound. To further improve DNN robustness, many advanced adversarial training methods have been proposed. TE \cite{dong2021exploring} improves SAT by mitigating a memorization issue. FGSM-SDI \cite{jia2022boosting} improves adversarial initialization. TRADES \cite{zhang2019theoretically} and MART \cite{wang2019improving} use loss regularization terms to make a trade-off between adversarial robustness and standard accuracy. DAT \cite{wang2019convergence} uses converge quality as a criterion to adjust adversarial training noises. ATES \cite{sitawarin2020improving} and CAT \cite{cai2018curriculum} apply curriculum strategy for adversarial training. Some other methods adjust the noise upper bound in the training process, including IAAT \cite{balaji2019instance}, FAT \cite{zhang2020attacks}, Customized AT \cite{cheng2020cat} and MMA \cite{ding2019mma}. GAIRAT \cite{zhang2020geometry} applies sample-wise weights to an adversarial training loss. LBGAT \cite{cui2021learnable} combines a teacher model and a student model for adversarial training.

However, training a model by those methods will significantly harm the model's standard accuracy (i.e., accuracy on clean data) \cite{tsipras2019robustness, raghunathan2019adversarial}, and removing the trade-off between robustness and accuracy is considered to be impossible \cite{zhang2019theoretically}. This issue prevents adversarial training from being used in many domains, such as medical image analysis, as practitioners do not want to sacrifice standard accuracy too much in exchange for robustness. 

Our study aims to lift the trade off between adversarial robustness and standard accuracy: alleviate (or even avoid) the reduction in standard accuracy while improving adversarial robustness. Our major innovations and contributions are: (1) We design a novel adversarial training method, named Increasing-Margin Adversarial (IMA) Training: In each training epoch, IMA makes sample-wise estimations of the upper bound of the adversarial noises, which will be used to generate adversarial samples in the next epoch. In the training process, decision boundaries will be gradually pushed away from the clean samples, which enhances robustness. Once an equilibrium state is reached, the adversarial noises will stop increasing, preventing adding too much noise that may hurt the model's standard accuracy. Extensive experiments are conducted to evaluate our method and the other eight representative defense methods by using AutoAttack \cite{croce2020reliable} and white-noise attack. The results show that: IMA significantly improves the adversarial robustness of DNNs and outperforms the other defense methods on the evaluated datasets and DNNs; IMA has the least degradation in standard accuracy among all the evaluated defense methods. (2) We apply IMA to the COVID-19 CT image classification application to show the significance of the robustness study in the medical domain. (3) We extend IMA for medical image segmentation and show that it can improve adversarial robustness with minimal reduction in standard accuracy for three applications. The result of prostate image segmentation application shows that both adversarial robustness and standard accuracy are improved. To our knowledge, for the first time, we demonstrate that it is possible to avoid the trade-off between standard accuracy and adversarial robustness for medical image segmentation.

\section{Methods}

\subsection{Terminology}

For simplicity, we use ``\textbf{noise level}" to refer to the upper bound $\epsilon$ of the adversarial noise for training or testing. ``\textbf{Standard accuracy}'' denotes a model's accuracy on clean data, which is also called clean accuracy in some literature. ``\textbf{Adversarial accuracy}'' denotes a model's accuracy on noisy/adversarial data, which measures the model's adversarial robustness and is called robust accuracy in some literature. A sample's ``\textbf{margin}" denotes the distance between the sample and the decision boundary of a classifier in the input data space. A ``\textbf{clean sample}" denotes a sample without adversarial noise. A ``\textbf{noisy sample}" denotes a noisy/adversarial sample.

\subsection{Adversarial Sample Generation by Projected Gradient Descent}
\label{2.1}
\begin{figure}[ht]
\centering
\begin{subfigure}{0.3\textwidth}
\includegraphics[width=0.98\linewidth]{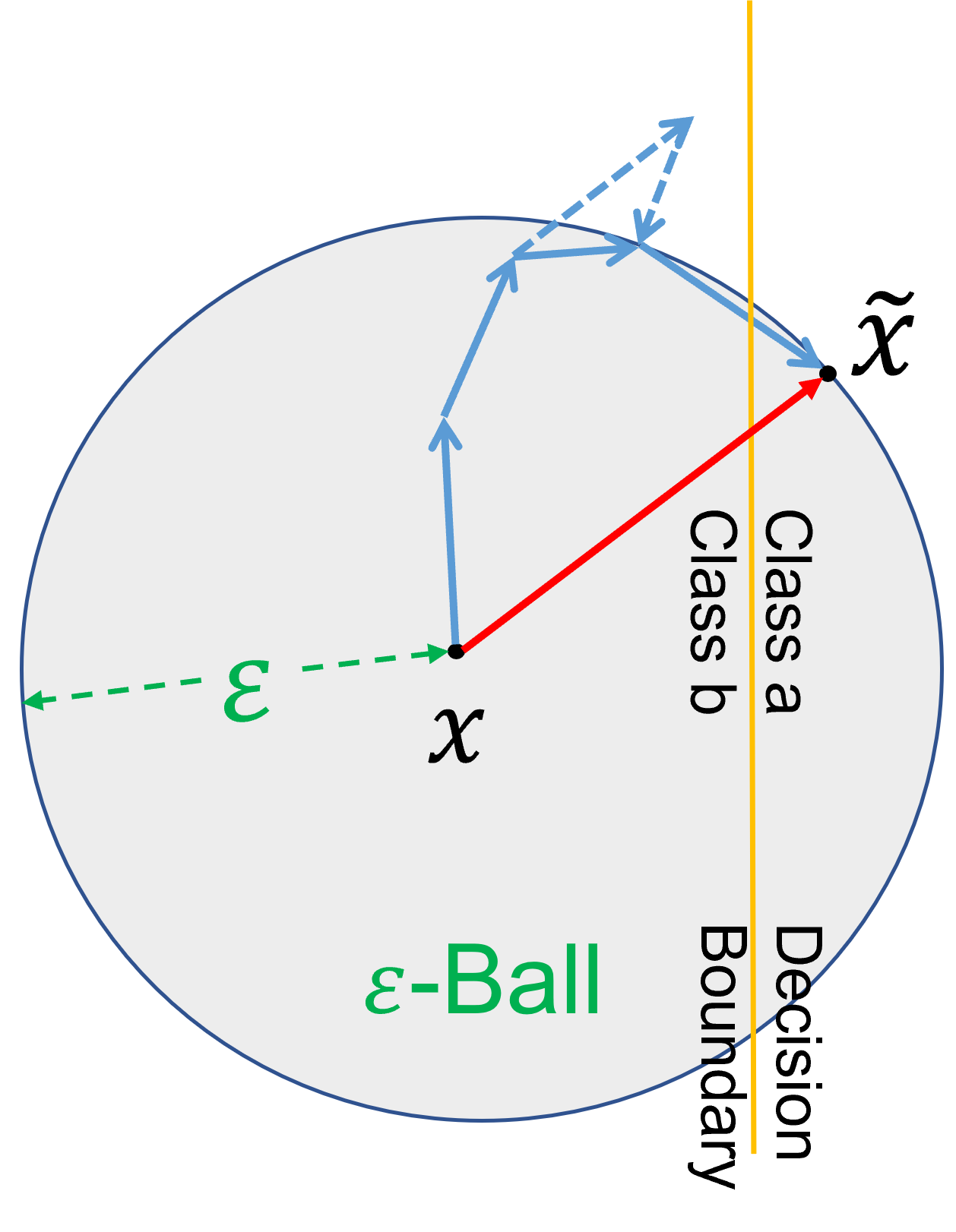}
\caption{}
\end{subfigure}
\begin{subfigure}{0.33\textwidth}
\includegraphics[width=0.98\linewidth]{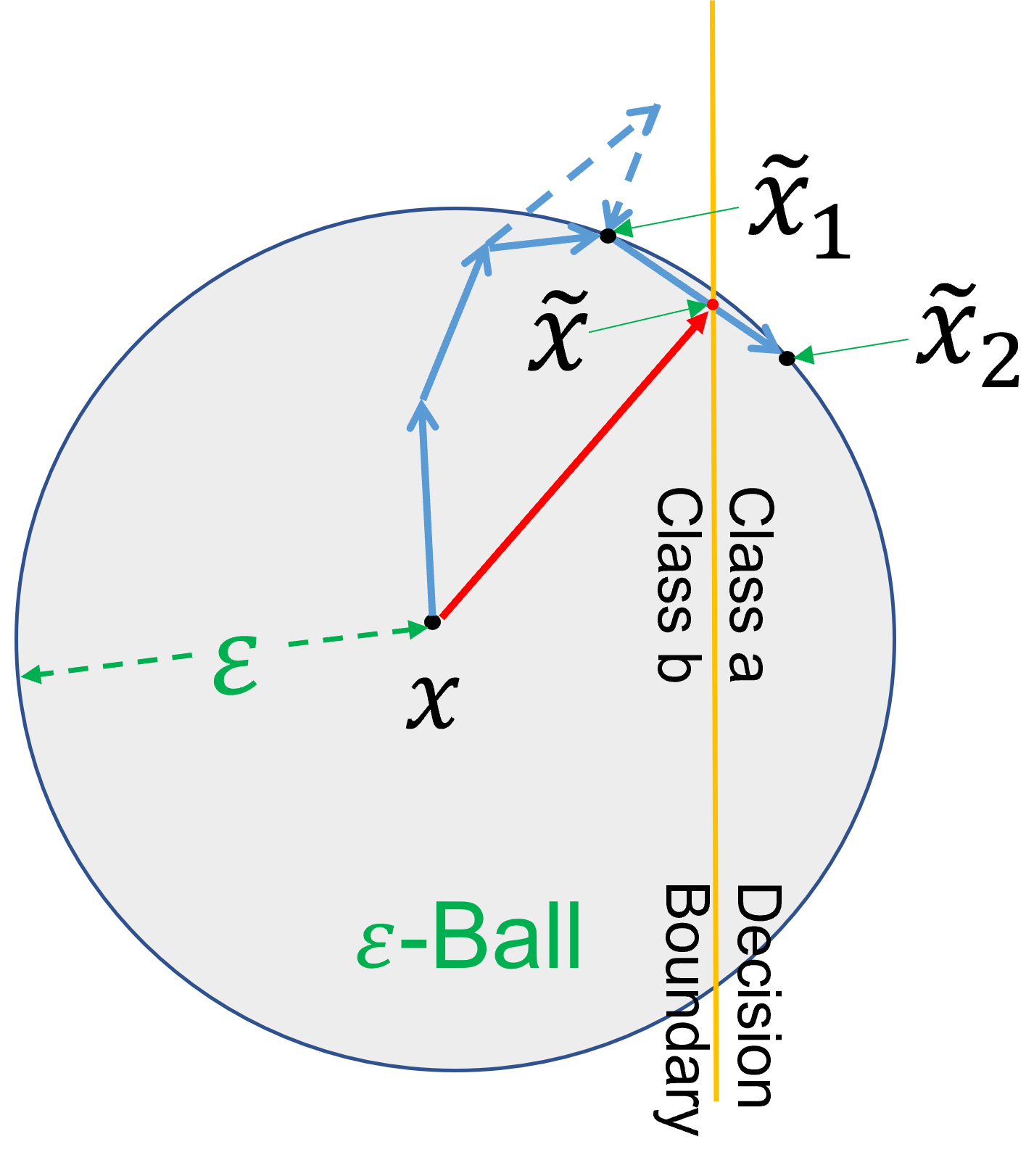}
\caption{}
\end{subfigure}
\caption{(a) Create an adversarial sample $\tilde{x}$ by PGD (Eq. (3)): a solid blue arrow shows one PGD iteration to update $\tilde{x}$; a dashed blue arrow shows that when $\tilde{x}$ moves out of the $\epsilon$-ball, it will be projected back onto the $\epsilon$-ball; the solid red arrow denotes the final adversarial noise. (b) Explanation of IMA Algorithm 2: during the PGD iterations (Eq. (3)), two types of adversarial samples are obtained, $\tilde{x}_1$ that is close to but does not cross the decision boundary, and $\tilde{x}_2$ that has just crossed the decision boundary. Then, a binary search is performed between $\tilde{x}_1$ and $\tilde{x}_2$ to locate the optimal $\tilde{x}$, which will be used for robust training. 
}
\label{fig2}
\end{figure}

 Let $(x,y)$ be a pair of a sample $x$ and its true label $y$. The objective function of generating an adversarial sample is: 
 \begin{equation}
\scriptsize
\tilde{x} = argmax_{x'} \; Loss(f_{\theta}(x'), y)
\end{equation}
where $Loss(.)$ is the loss function, $\tilde{x}$ is the adversarial sample to be generated, and $f_{\theta}(.)$ is a DNN model with parameter $\theta$. The objective function is under the constraint that:
\begin{equation}
\scriptsize
 \tilde{x} \in \{x' | \; \|x - x'\|_p \le \epsilon\}
\end{equation}
where $\epsilon$ is noise upper bound and $\| . \|_p$ is vector Lp-norm. 

Given a clean sample $(x,y)$, there are many ways to solve Eq.(1) to obtain an adversarial sample $\tilde{x}$. One method is called Projected Gradient Descent (PGD) \cite{madry2017towards, kurakin2016adversarial}, which leverages an iterative way to solve Eq.(1):
\begin{equation}
\scriptsize
 x^{(k)} \gets \Pi_{\epsilon} (\alpha \cdot h( \nabla_{x^{(k-1)}} Loss(f_{\theta}(x^{(k-1)}), y) + x^{(k-1)} ) 
\end{equation}
where $h(.)$ is the normalization function, $\alpha$ is the step size, $x^{(k)}$ is the adversarial sample at the iteration $k$,  $\Pi_{\epsilon} (.)$ is a projection operation to ensure the generated noise is within an $\epsilon$-ball (i.e., $\| x^{(k)}-x \|_p \le \epsilon$), and $\nabla$ is the gradient operator. After K iterations, the adversarial sample is obtained: $\tilde{x}$=$x^{(K)}$, which leads to a significant increase in the loss. For a DNN model, a large loss increase leads to a wrong output, given $\tilde{x}$ as the input.

\subsection{Standard Adversarial Training (SAT)}
\label{2.2}
By adding adversarial noises to the training samples, the Standard Adversarial Training (SAT) \cite{madry2017towards} has the following objective function:
\begin{equation}
\scriptsize
 min_{\theta} \; Loss(f_{\theta}(\tilde{x}), y)
\end{equation}
where $\tilde{x}$ is an adversarial sampled generated by the PGD method (Eq.(3)) with the constrain $\|x' -x\|_p \le \epsilon_{train}$. Here, the training noise level $\epsilon_{train}$ is the same for every clean sample $\{x,y\}$ in the training set.

The SAT uses adversarial samples $\{\tilde{x},y\}$ to train the DNN model $f_{\theta}$. In this way, the DNN model may become robust against adversarial noises. Using a fixed and uniform noise level (i.e., noise upper bound) $\epsilon_{train}$ for all training samples is the weak point of SAT.

\begin{figure}[ht]
\centering
\begin{subfigure}{0.48\textwidth}
\includegraphics[width=0.98\linewidth]{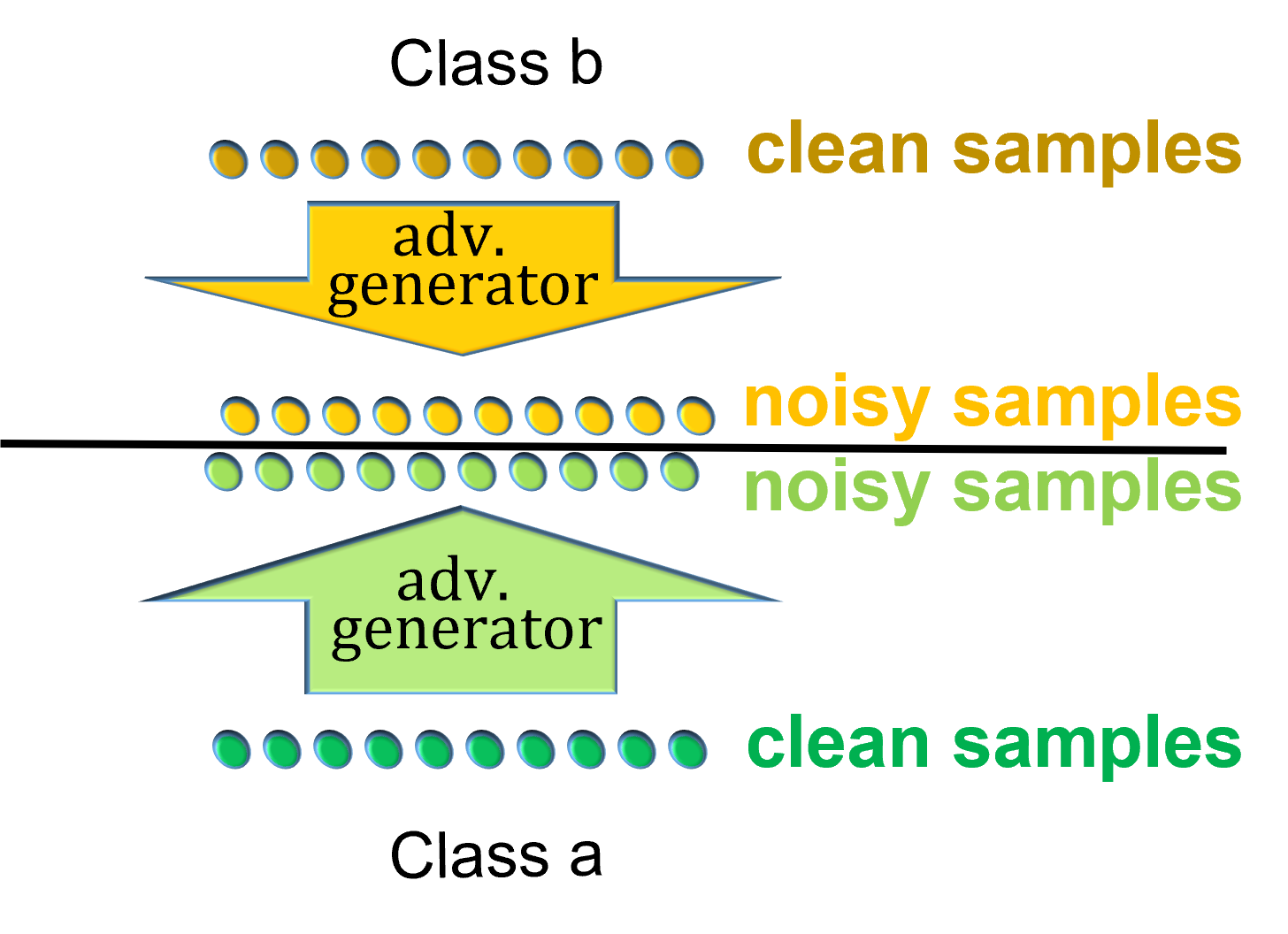}
\caption{}
\end{subfigure}
\begin{subfigure}{0.48\textwidth}
\includegraphics[width=0.98\linewidth]{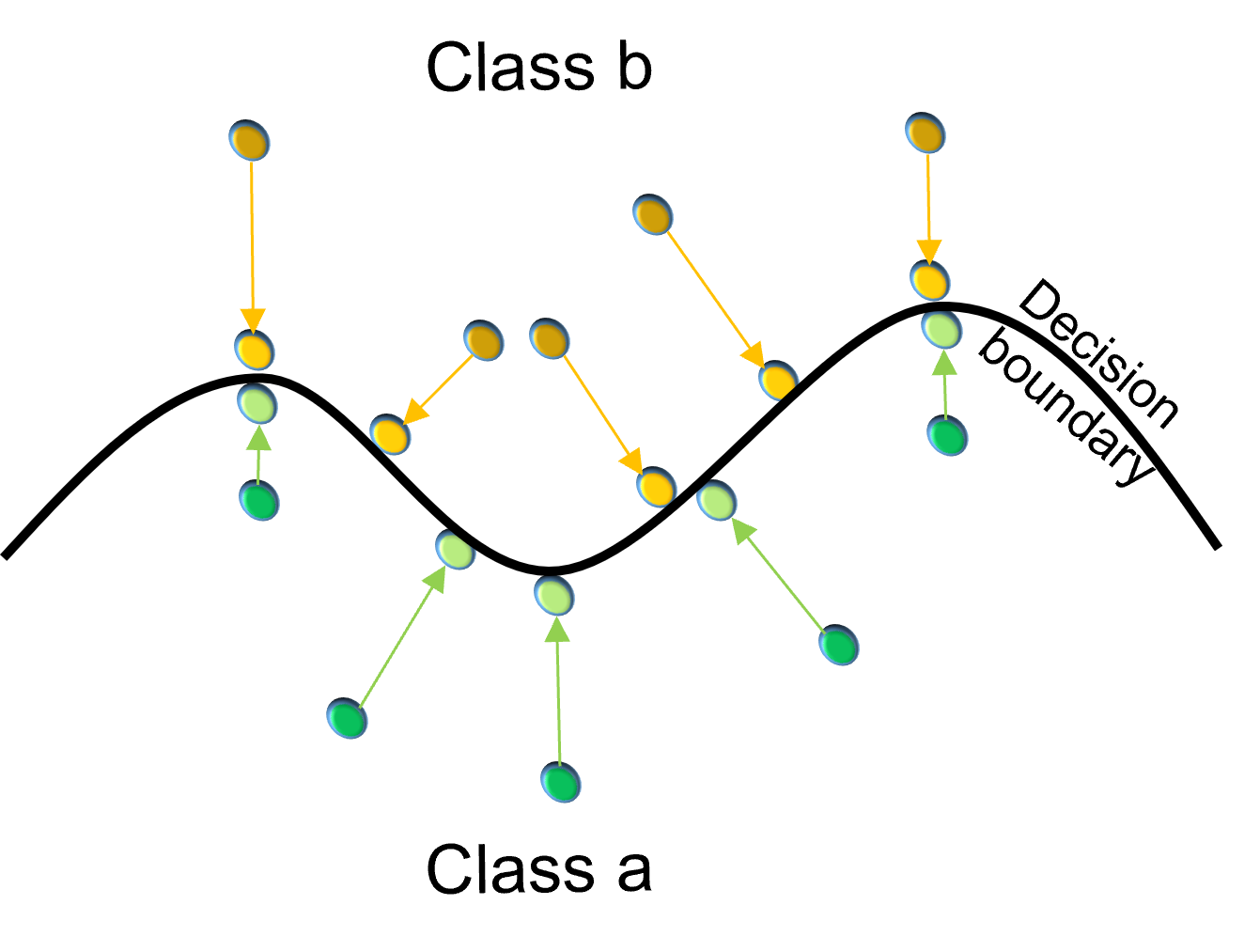}
\caption{}
\end{subfigure}
\caption{(a): To improve robustness, ideally, adversarial training samples should be placed/generated closer to the true/optimal decision boundary. By training the model with those optimal adversarial samples, the model decision boundaries will be pushed away from the clean samples, and therefore adversarial robustness is improved. (b): However, the distribution of the samples and the optimal decision boundary are not always as ideal as that in (a). Given a complex sample distribution and a nonlinear decision boundary, the margin of each (clean) sample is different. A uniform and fixed adversarial training noise can lead to adversarial training samples that go across the true decision boundary, and a model trained on these adversarial samples will have low standard accuracy on clean data. Apparently, the optimal adversarial training samples should be close to the decision boundary. Training with too small noise is not effective enough. Training with too large noise leads to low standard accuracy on clean data.
}
\label{fig3}
\end{figure}

\subsection{Our Method: Increasing-Margin Adversarial (IMA) Training}
\label{2.4}


Before introducing the algorithms of our method, we use simple examples to illustrate the basic idea. From Fig. \ref{fig3}, the optimal locations of the adversarial training samples $\tilde{X}$ should be very close to the decision boundary (Fig. \ref{fig3} (b)). And the optimal adversarial noise level $\epsilon_{train}$ should be the distance between the (true) decision boundary and each sample $x$. This distance is similar to the ``Margin'' in Support Vector Machine (SVM) \cite{cortes1995support}. So, we will call this optimal adversarial noise upper bound as ``Margin'' in this paper. This margin should be the shortest distance between $x$ and the corresponding decision boundary, and an optimal $\tilde{x}$ should just be about to cross the decision boundary. We refer the reader to the Appendix A for a more rigorous analysis about optimal adversarial training samples.

\begin{algorithm}[ht]
\scriptsize
    \caption{IMA Training in One Epoch}
    \label{alg1}

    \textbf {Input:} \\
     the training set $S$ \\
     the DNN model $f(.)$ \\
     $g(.)$ is the function that transforms the output of $f(.)$ to a predicted class label, e.g., $argmax$\\
     $Loss$ is the loss function for training the model $f$.\\
    $\mathcal{E}$ is the array of the estimated sample margins. $\mathcal{E}(i)$ is the margin of the sample with the unique ID $i$. Every $\mathcal{E}(i)$ is initialized to be $\Delta_\epsilon$\\
    \textbf {Parameters:} \\
    $\Delta_\epsilon$ is the expansion step size (a positive scalar) \\
    $\epsilon_{train}$ is the adversarial training noise upper bound\\
  \textbf{Output:} Updated model $f$ after this training epoch\\
  \textbf{Process:} 

    \begin{algorithmic}[1]
    \FOR {each training sample ($x$, $y$) with ID $i$ in $S$}   
    \STATE Run the model $f$ on the clean sample: $z \gets f(x)$ 
    \STATE  $L_0 \gets Loss(z,\ y)$  
    \IF {$g(z) = y$}
    \STATE Generate a noisy sample using the \textbf{Algorithm 2}: $\tilde{x}, \beta \gets GenAdv(x, y,\mathcal{E}(i),f, Loss)$ 
    \STATE Run the model $f$ on the noisy sample: ${\tilde{z}}\gets f(\tilde{x})$\\
    \STATE  $L_1 \gets Loss(\tilde{z},\ y)$
    \ELSE
    \STATE $L_1 \gets L_0$    
    \ENDIF  
    \STATE  $L_2 \gets (L_0+ L_1)/2 $
    \STATE  Back-propagate from the combined loss $L_2$ and update the model $f(.)$  
	\IF{$g(\tilde{z}) = y $ \textbf{and} $\beta = 0$ }		
		\STATE $\mathcal{E}(i) \gets \mathcal{E}(i)+\Delta_\epsilon$  (Enlarge the margin)
	\ELSIF {$g(\tilde{z})\ != y $ \textbf{and} $\beta = 1$ }	
		\STATE $\mathcal{E}(i) \gets (||x-\tilde{x}||_p + \mathcal{E}(i))/2$  (Shrink)
	\ENDIF
    \ENDFOR \\ 
    \STATE Clip every element of $\mathcal{E}$ into the range of $[0, \epsilon_{train}]$	\\
    \end{algorithmic} 
\textbf{Note:} 
This algorithm is implemented in mini-batches. $||.||_p$ denotes vector Lp norm. 	
\end{algorithm}

We design IMA to preserve standard accuracy as much as possible while improving robustness. One epoch of the IMA training process is shown in Algorithm \ref{alg1}, which includes two sub-processes: (1) Compute the loss and update the DNN model (Line 2 to 12); (2) Update the sample margin estimation (Line 13 to 17). Here is a brief description of Algorithm \ref{alg1}. $x$ is a clean training sample with true label $y$, and $i$ is the unique ID for this sample (Line 1). After the clean sample is processed by the DNN model $f(.)$, the loss $L_0$ on the clean sample is obtained (Line 2-3). If $x$ is correctly classified (Line 4), Algorithm \ref{alg2} will be used to generate an adversarial/noisy sample $\tilde{x}$ (Line 5). After the noisy sample is processed by the model, the loss $L_1$ on the noisy sample (Line 6-7) is obtained. If $x$ is misclassified, Algorithm \ref{alg2} will be skipped (Line 8-9). Then, the model  $f(.)$ is updated by backpropagation from the combined loss (Line 11-12). If both the clean sample $x$ and the noisy sample $\tilde{x}$ are correctly classified, which means the estimated sample margin is not large enough to reach the current decision boundary, the estimated margin $\mathcal{E}(i)$ for this training sample will be expanded (Line 13-14). Otherwise, the estimated sample margin $\mathcal{E}(i)$ is too large and the noise with this magnitude has already pushed $\tilde{x}$ across the current decision boundary, and therefore, $\mathcal{E}(i)$ should be reduced (Line 16). $\mathcal{E}(i)$ should always be smaller than the maximum noise $\epsilon_{train}$, ensured by a clip operation (Line 19).

\begin{algorithm}[H]
\scriptsize
    \caption{(GenAdv): generate adversarial/noisy samples}
    \label{alg2}
    \textbf {Input:} \\
    training sample ($x$, $y$) with ID $i$\\
    the estimated margin $\mathcal{E}(i)$, currently\\
    the DNN model $f(.)$\\
    $g(.)$ is the function that transforms the output of $f(.)$ to a predicted class label, e.g., $argmax$\\
    the Loss function $L$ \\
  \textbf {Parameters:} \\
  maximum iteration number $K \gets 20$ \\
  step size $\alpha \gets (4 \times \mathcal{E}(i)) / K$ \\
  \textbf{Output:} the generated noisy sample $\tilde{x}$ and the binary indicator $\beta$ \\ 
  \textbf{Function} GenAdv($x$, $y$, $\mathcal{E}(i)$, $f$, $L$): 
    \begin{algorithmic}[1]   
    \STATE $\beta \gets 0$
    \WHILE {$K > 0$}   
    \STATE $\tilde{x} \gets \Pi_{\mathcal{E}(i)} (\alpha \cdot h(\nabla_{x} L(f(x),y)) + x ) $
    \IF {$g(f(\tilde{x})) \neq y$ }
    \STATE $\tilde{x} \gets$ BinarySearch ($\tilde{x}$, $x$, $f$, $y$)
    \STATE $\beta \gets 1$
    \STATE \textbf{return} $\tilde{x}$,$\beta$
    \ENDIF
	\STATE $x \gets \tilde{x}$	
    \STATE $K \gets K-1$
    \ENDWHILE 
    \STATE \textbf{return} $\tilde{x}$, $\beta$
    \end{algorithmic} 
\textbf{Function} BinarySearch($\tilde{x}_2$, $\tilde{x}_1$, $f$, $y$):
\begin{algorithmic}[1] 
\STATE $N \gets 10$
\STATE $\tilde{x} \gets (\tilde{x}_1+\tilde{x}_2)/2$
\WHILE {$N > 0$}
\IF {$g(f(\tilde{x} )) \neq y $}
\STATE $\tilde{x}_2 \gets \tilde{x} $
\ELSE
\STATE $\tilde{x}_1 \gets \tilde{x} $
\ENDIF
 \STATE $N \gets N-1$
\STATE $\tilde{x} \gets (\tilde{x}_1+\tilde{x}_2)/2$
 \ENDWHILE 
 \STATE \textbf{return} $\tilde{x}$
\end{algorithmic} 
\textbf{Note:} 
This algorithm is implemented in mini-batches. $h(.)$ is the normalization function. $||.||_p$ denotes vector Lp norm. $N$ is always set to $10$, which is constant. So, this binary search will not enlarge the time complexity of the whole algorithm.
$\Pi_{\epsilon} (.)$ ensures that $||\tilde{x}-x||_p \le \epsilon$.
\end{algorithm}

Here is a brief description of Algorithm \ref{alg2} (GenAdv). In each iteration, noise is added to $x$ to get $\tilde{x}$ (Line 3, see Eq. (3)). If $\tilde{x}$ is misclassified by the model $f(.)$ (Line 3), then $\tilde{x}$ is represented by $\tilde{x}_2$, and $x$ is replaced by $\tilde{x}_1$ in Fig. \ref{fig2}. So, a binary search is applied to find the new $\tilde{x}$ (Line 5, Fig. \ref{fig2}), which is just about to cross the decision boundary.  If the misclassification (Line 4) does not happen, then the algorithm runs to the next iteration.


\subsection{Modification of Algorithm \ref{alg2} to Work With Data Augmentation}
\label{2.6}

Data augmentation (e.g., flip an image) is widely used to improve DNN standard accuracy on clean data (e.g., CIFAR10). This causes a problem for Algorithm 2: the optimal $\tilde{x}$ may not be between $\tilde{x}_1$ and $\tilde{x}_2$. Therefore, in the case of data augmentation, the function GenAdv in Algorithm \ref{alg2} is replaced with Algorithm \ref{alg3}, in which the binary search is conducted between $x$ and $\tilde{x}$.

\begin{algorithm}[H]
\scriptsize
    \caption{GenAdv in case of Data Augmentation}
    \label{alg3}
  \textbf{Function} GenAdv($x$, $y$, $\mathcal{E}(i)$, $f$, $L$): 
    \begin{algorithmic}[1]    
    \STATE $x_{init} \gets x$
    \WHILE {$K > 0$}   
    \STATE $\tilde{x} \gets \Pi_{\mathcal{E}(i)} (\alpha \cdot h(\nabla_{x}L(f(x),y)) + x ) $
	\STATE $x \gets \tilde{x} $	
    \STATE $K \gets K-1$
    \ENDWHILE 
    \STATE \textbf{return}  BinarySearch ($\tilde{x}$, $x_{init}$, $f$, $y$)
    \end{algorithmic} 
\end{algorithm}

\subsection{Comparison with Related Work}
\label{2.7}


In this section, we briefly discuss the difference between our IMA and the other related defense methods \cite{dong2021exploring, zhang2019theoretically,  balaji2019instance, zhang2020attacks, ding2019mma, zhang2020geometry, cui2021learnable}, which have open-source implementations and are evaluated in our experiments.

TE \cite{dong2021exploring} improves SAT by mitigating a memorization issue. TRADES \cite{zhang2019theoretically} optimizes a loss with a KL divergence-based regularization term to make a trade-off between adversarial robustness and standard accuracy. 
GAIRAT \cite{zhang2020geometry} applies sample-wise weights in the loss, but the weights cannot prevent adversarial training samples from crossing the decision boundary. So, GAIRAT will likely generate adversarial training samples with too much noise. As shown in the experiments, GAIRAT's standard accuracy is much worse than that of IMA (see Table \ref{CIFARClassification} and Table \ref{MedicalClassification}). 
LBGAT \cite{cui2021learnable} needs a teacher model during adversarial training. 
The FAT method \cite{zhang2020attacks} applies early-stop PGD to generate adversarial training samples with a parameter $\tau$. When $\tau=0$, the generated adversarial training sample is essentially $\tilde{x}_1$ in Fig. \ref{fig2} (b); when $\tau>0$, the generated adversarial training sample goes across the decision boundary, which will hurt the model's standard accuracy and is against the idea of IMA. 
The IAAT method \cite{balaji2019instance} uses adaptive sample-wise adversarial training noises to train a model, but the sample-wise noises are based on heuristics, which may not be optimal.
The MMA method \cite{ding2019mma} heavily relies on the soft logit margin loss, which may not generate optimal adversarial samples.
Our method is obviously different from these methods.

\section{Results}

In this section, we report the approach for method evaluation and comparison, the configuration of image classification and segmentation experiments, and the results. Pytorch 1.9.0 \cite{paszke2017automatic} is used for implementation. Nvidia V100 GPUs are used for model training and testing.

\subsection{Method Evaluation}
We use AutoAttack \cite{croce2020reliable} to evaluate the adversarial robustness of DNNs that classify an entire image into different categories. AutoAttack consists of four attacks: AutoPGD (a white-box untargeted attack, stronger than the vanilla PGD), APGD-t (a white-box targeted attack), FAB-t (a white-box targeted attack), and Square (a black-box attack). AutoAttack is parameter-free, so there is no need to configure it manually.  

Dice loss is not supported in the official AutoAttack code \cite{croce2020reliable}, but Dice loss is the major loss for training image segmentation DNNs (e.g., nnUnet \cite{isensee2021nnu} used in our experiments). The authors of \cite{daza2021towards} modified AutoAttack to be used for segmentation tasks. So, we use the modified AutoAttack from \cite{daza2021towards} in our image segmentation experiments.

In addition to the evaluation using adversarial noises (measured by L2 norm and L-inf norm), we also use white noises to evaluate a model's robustness. The white noises are from a uniform distribution with noise upper bound measured by L-inf or L2 norm. Because white noises are highly randomized, we use white noises to attack each model with 100 iterations to get a stable result. This evaluation will show that an adversarially-robust DNN model is also robust to random noises \cite{fawzi2016robustness} that are everywhere in the real world. 



\subsection{Method Comparison for Image Classification Experiments}

In the experiments, a model trained only on clean data is named ``STD'' (i.e., standard training); a model trained by an adversarial training method is named by the method. For example, a model trained by SAT \cite{madry2017towards} is named ``SAT". Other methods include: FAT \cite{zhang2020attacks}, TRADES \cite{zhang2019theoretically}, GAIRAT\cite{zhang2020geometry}, MMA\cite{ding2019mma}, IAAT\cite{balaji2019instance}, TE \cite{dong2021exploring} and LBGAT \cite{cui2021learnable}.

\subsection{Method Comparison for Image Segmentation Experiments}
All the competing defense methods are not designed for image segmentation and therefore cannot be directly applied to our segmentation applications. We tried our best to modify some of the methods for image segmentation, which are SAT, TE, and TRADES. The MMA method \cite{ding2019mma} can hardly be modified because the algorithm and theory of MMA heavily rely on the soft logit margin loss proposed in \cite{ding2019mma}, and changing this loss to Dice loss will invalidate the optimality condition of the theory, and the zero-crossing search in the method does not apply for Dice loss (it is always nonnegative). The performance of LBGAT \cite{cui2021learnable} relies on the teacher model; but the original method \cite{cui2021learnable} only provides a teacher model for classification tasks; thus, we would need to design a new teacher model for image segmentation tasks. The other three defense methods, GAIRAT \cite{zhang2020geometry}, IAAT \cite{balaji2019instance}, and FAT \cite{zhang2020attacks} use some criteria in adversarial training, which are not clearly defined for segmentation tasks. To use these methods for the segmentation tasks, significant modifications to their algorithms are needed, which is analog to changing the engine of a Ferrari, and the modified car can no longer be called Ferrari. So, these three defense methods are unsuitable for comparison in the segmentation tasks.

As a result, for the image segmentation experiments, we only compare IMA with TRADES, TE, SAT, and STD, by replacing the cross-entropy loss with dice loss that is used in nnUnet \cite{isensee2021nnu}.

To make a fair comparison, the same training noise level $\epsilon_{train}$ is used for all the defense methods in each of the experiments. In the CIFAR10 image classification experiment, the $\epsilon_{train}$ is 3, to be consistent with that in the paper \cite{ding2019mma}. Then, we calculate the average pixel-wise noise level $\epsilon_p$ (for training, measured by L2 norm):
\begin{equation}
\scriptsize
\sqrt[2]{32 \times 32 \times 3 \times \epsilon_p} = 3\\
\end{equation}
Thus, $\epsilon_p = \frac{3}{32 \times 32}$. We apply this $\epsilon_p$ to PathMNIST (28$\times$28$\times$3 image size) and COVID-19 (224$\times$224$\times$1 image size) datasets to obtain the training noise level $\epsilon_{train}$. For the PathMNIST experiment, $\epsilon_{train} = \sqrt[2]{28 \times 28 \times 3 \times \epsilon_p} \approx 3$. For the COVID-19 experiment $\epsilon_{train} = \sqrt[2]{224 \times 224 \times 1 \times \epsilon_p} \approx 12$. For each of the image segmentation experiments, we also make sure the same training noise level is used for all the defense methods. 


\subsection{Configuration for Image Classification Experiments}
\label{3.1}

\subsubsection{Configuration for CIFAR10 Image Classification} 
\label{3.2.3}

CIFAR10 dataset \cite{krizhevsky2009learning} contains 60000 color images in 10 classes. The image size is 32$\times$32$\times$3. Each class has 6000 images, 5000 for training, and 1000 for testing. We apply all the methods to WideResNet-28-4 (WRN-28-4) that is  widely used for method evaluation \cite{he2016deep, ding2019mma}. Data augmentation is used during training. For all methods, the batch size is 128. The training noise level $\epsilon_{train}$ is 3.0. A grid search is done to find an optimal $\Delta_\epsilon$ in IMA.  The $\Delta_\epsilon$ is 0.03 (see \ref{A1}). IMA is trained for 150 epochs, which is about the same number of epochs used in \cite{ding2019mma}. The optimizer is SGD with the same parameters as those of \cite{ding2019mma}. The configurations of the other methods are consistent with those in their papers.

\subsubsection{Configuration for Colon Pathology Image Classification} 

The PathMNIST dataset \cite{medmnistv2} is collected for predicting survival from colorectal cancer histology slides, which has a training dataset of 100,000 non-overlapping image patches from hematoxylin \& eosin stained histological images (NCT-CRC-HE-100K), and a test dataset of 7,180 image patches from a different clinical center (CRC-VAL-HE-7K). The dataset comprises 9 types of tissues, resulting in a multi-class classification task. The source images of 3$\times$224$\times$224 are resized into the size of 3$\times$28$\times$28. NCT-CRC-HE-100K is split into training and validation sets with a ratio of 9 : 1. The CRC-VAL-HE-7K is treated as the test set. We apply all the methods to ResNet-18 used in \cite{medmnistv2}. Data augmentation is used during training. For all methods, the training noise level $\epsilon_{train}$ is 3.0. The $\Delta_\epsilon$ in IMA is 0.04 (see \ref{A1}). The number of training epochs and optimizer settings are the same as those in Section \ref{3.2.3}. The configurations of the other methods are consistent with those in their papers.

\subsubsection{Configuration for COVID-19 CT Image Classification}

 We used a public COVID-19 CT image dataset \cite{soares2020sars}. It was collected from patients in hospitals in San Paulo, Brazil. It contains 1252 CT scans (2D images) from COVID-19-infected patients and 1230 CT scans (2D images) from uninfected patients. From the infected cases, we randomly selected 200 samples for testing, 30 for validation, and 1022 for training. From the uninfected cases, we randomly selected 200 for testing, 30 for validation, and 1000 for training. These images have different sizes. To facilitate image classification, each image is resized to $224\times224$, which is a standard procedure in machine learning. We modified the output layer of the Resnet-18 model \cite{he2016deep} for this binary classification task: uninfected (label 0) vs infected (label 1). We also replaced batch normalization with instance normalization because it is known that batch normalization is unstable for small batch-size \cite{wu2018group}. As shown in the previous studies \cite{shi2020review}, infected regions in the images have a special pattern called ground-glass opacity. For all methods, the batch size is 32, the number of training epochs is 100, and the training noise level $\epsilon$ is 12. The $\Delta_\epsilon$ in IMA is 2.0 (see \ref{A1}). The optimizer settings are the same as those in Section \ref{3.2.3}. Other configurations of the other methods are consistent with those in their papers.

\subsection{Configuration for Image Segmentation Experiments}

\subsubsection{The DNN - nnUnet}

We use nnUnet \cite{isensee2021nnu}, a well-known DNN for medical image segmentation. The nnUnet can automatically configure itself, including preprocessing, network architecture, training, and post-processing for the dataset. The inputs of nnUnet are 2D slices of 3D images. Dice score is used to measure segmentation accuracy, which is in the range of 0 to 1.

\subsubsection{Extension of IMA for Image Segmentation}

In this experiment, we show that IMA can improve the adversarial robustness of nnUnet for clinical medical image segmentation. To apply IMA to image segmentation, we need to define the ``correctness" of a segmentation result. Since the Dice index is often used to evaluate segmentation performance, a segmentation result can be considered ``correct" if Dice $>$ threshold, and ``wrong" otherwise, which is a binary classification to classify the segmentation result. In the experiment, this Dice threshold is set to 60\% for all of the datasets because a Dice score higher than 60\% is considered ``good'' for many medical applications \cite{visser2020accurate, visser2019inter, cicchetti1994guidelines, bartko1991measurement}.

\subsubsection{Configuration for Heart MRI Image Segmentation}

The Heart MRI dataset \cite{simpson2019large} has 20 labeled 3D images: 16 for training, 1 for validation, and 3 for testing. The median shape of each 3D image is 115 $\times$ 320 $\times$ 320, of which 115 is the number of slices. In this experiment, only 2D segmentation is considered, so the nnUnet model's input is one slice. The batch size (40), and input image size (320 $\times$ 256) are self-configured by nnUnet for this dataset. The model is trained for 50 epochs. Other training settings are documented in \cite{isensee2021nnu}. IMA step size $\Delta_\epsilon$ is 5 (see Appendix B). $\epsilon_{train}$ is 20 for all methods.

\subsubsection{Configuration for Hippocampus MRI Image Segmentation}

The Hippocampus MRI dataset \cite{simpson2019large} has 260 labeled 3D images: 208 for training, 17 for validation, and 35 for testing. The median shape of each 3D image is 36 $\times$ 50 $\times$ 35, where 36 is the number of slices. The batch size (366), the input image size (56 $\times$ 40), and the network structure are self-configured by nnUnet for this dataset. The model is trained for 50 epochs, where each epoch has 50 iterations. Other training settings are the same as those in \cite{isensee2021nnu}. IMA step size $\Delta_\epsilon$ is 2 (see Appendix B). $\epsilon_{train}$ is 15 for all methods.

\subsubsection{Configuration for Prostate MRI Image Segmentation}

The Prostate MRI dataset \cite{simpson2019large} has 32 labeled 3D images: 25 for training, 2 for validation, and 5 for testing. The median shape of each 3D image is $20 \times 320 \times\ 319$, where 20 is the number of slices. The batch size (32), the input image size ($320 \times 320$), and the network structure are self-configured by nnUnet for this dataset. The model is trained for 50 epochs, where each epoch has 50 iterations. Other training settings are the same as those in \cite{isensee2021nnu}. IMA step size $\Delta_\epsilon$ is 10 (see Appendix B). $\epsilon_{train}$ is 40 for all methods.

\subsection{Experiment Results}
\label{3.2}

\begin{table}[H]
    \caption{Results on CIFAR10. The metric is classification accuracy (\%), and the largest value in each column is bold. AutoAttack and white-noise attack are performed using two norm settings: L-inf and L2 norm to measure noise level.}
    \label{CIFARClassification}
    \begin{subtable}{.49\linewidth}
      \centering
        \caption{AutoAttack on CIFAR10 (L-inf)}
        \label{CIFAR101}
\resizebox{0.95\columnwidth}{!}{
  \begin{tabular}{|l|lllll|}
    \toprule
   Noise level & 0&2/255&4/255&8/255&Avg.\\
    \midrule
STD	&94.2	&0	&0	&0	&23.55\\
    \midrule
IMA (Ours)	&\textbf{87.89}	&\textbf{76.28}	&61.89	&34.27	&\textbf{65.08}\\
MMA \cite{ding2019mma}	&82.11	&71.68	&59.39	&35.38	&62.14\\
GAIRAT\cite{zhang2020geometry}	&74.40	&61.07	&46.76	&24.58	&51.70\\
FAT\cite{zhang2020attacks}	&82.45	&72.90	&\textbf{62.02}	&\textbf{37.49}	&63.71\\
TRADES\cite{zhang2019theoretically}	&76.84	&57.98	&40.23	&19.49	&48.63\\
IAAT\cite{balaji2019instance}	&83.26	&72.43	&59.73	&33.25	&62.17\\
LBGAT\cite{cui2021learnable}	&68.45	&59.87	&50.85	&33.29	&53.12\\
TE \cite{dong2021exploring}	&58.29	&51.49	&46.01	&33.24	&47.25\\
SAT\cite{madry2017towards}	&61.64	&55.10	&48.77	&36.70	&50.55\\
    \bottomrule
  \end{tabular}
  }
    \end{subtable}%
    \begin{subtable}{.49\linewidth}
      \centering
        \caption{White noise on CIFAR10 (L-inf)}
        \label{CIFAR102}
        \resizebox{.95\columnwidth}{!}{
  \begin{tabular}{|l|lllll|}
    \toprule
   Noise level & 0& 8/255& 16/255& 32/255&Avg.\\
    \midrule
STD	&94.2	&86.97	&76.14	&50.03	&76.83\\
    \midrule
IMA (Ours)	&\textbf{87.89}	&\textbf{87.17}	&\textbf{86.40}	&\textbf{84.10}	&\textbf{86.39}\\
MMA \cite{ding2019mma}	&82.11	&81.38	&80.64	&78.37	&80.62\\
GAIRAT\cite{zhang2020geometry}	&74.40	&69.07	&63.31	&69.12	&68.98\\
FAT\cite{zhang2020attacks}	&82.45	&81.54	&80.59	&78.75	&80.83\\
TRADES\cite{zhang2019theoretically}	&76.84	&75.61	&73.96	&70.02	&74.11\\
IAAT\cite{balaji2019instance}	&83.26	&82.75	&81.77	&79.65	&81.85\\
LBGAT\cite{cui2021learnable}	&68.45	&67.67	&66.84	&64.46	&66.85\\
TE \cite{dong2021exploring}	&58.29	&57.76	&57.17	&55.92	&57.29\\
SAT\cite{madry2017towards}	&61.64	&61.08	&60.44	&58.68	&60.46\\
    \bottomrule
  \end{tabular}
        }
    \end{subtable}  
    \begin{subtable}{.49\linewidth}
      \centering
        \caption{AutoAttack on CIFAR10 (L2)}
        \label{CIFAR103}
\resizebox{0.95\columnwidth}{!}{
  \begin{tabular}{|l|lllll|}
    \toprule
   Noise level & 0&0.3&0.6&0.9&Avg.\\
    \midrule
STD	&94.2	&0	&0	&0	&23.55\\
    \midrule
IMA (Ours)	&\textbf{87.89}	&\textbf{75.67}	&\textbf{60.66}&	45.27	&\textbf{67.37}\\
MMA \cite{ding2019mma}	&82.11	&71.66	&59.35	&\textbf{46.93}	&65.01\\
GAIRAT\cite{zhang2020geometry}	&74.40	&52.60	&31.70	&16.00&	43.68\\
FAT\cite{zhang2020attacks}	&82.45	&67.65	&49.28	&31.21	&57.65\\
TRADES\cite{zhang2019theoretically}&	76.84	&56.59	&38.23	&26.18	&49.46\\
IAAT\cite{balaji2019instance}	&83.26	&72.77	&60.27	&46.47	&65.69\\
LBGAT\cite{cui2021learnable}	&68.45	&60.07	&51.56	&42.53	&55.65\\
TE \cite{dong2021exploring}	&58.29	&52.55	&46.75	&41.24	&49.71\\
SAT\cite{madry2017towards}	&61.46	&55.60	&49.78	&43.97	&52.70\\
    \bottomrule
  \end{tabular}
  }
    \end{subtable}%
    \begin{subtable}{.49\linewidth}
      \centering
        \caption{White noise on CIFAR10 (L2)}
        \label{CIFAR104}
        \resizebox{.95\columnwidth}{!}{
  \begin{tabular}{|l|lllll|}
    \toprule
   Noise level & 0& 1& 2& 3&Avg.\\
    \midrule
STD	&94.2	&86.86&	76.29&	63.92&80.32\\
    \midrule
IMA (Ours)	&\textbf{87.89}	&\textbf{87.16}	&\textbf{86.32}	&\textbf{85.41} &\textbf{86.70}\\
MMA \cite{ding2019mma}	&82.11	&81.46	&80.67&	79.62&80.97\\
GAIRAT\cite{zhang2020geometry}	&74.40&	73.00&	71.92&	70.63&72.49\\
FAT\cite{zhang2020attacks}	&82.45	&81.49	&80.68	&79.76 &81.10\\
TRADES\cite{zhang2019theoretically}	&76.84	&75.62	&73.99&	72.24&74.67\\
IAAT\cite{balaji2019instance}	&83.26	&82.27&	81.86	&80.83 &82.05\\
LBGAT\cite{cui2021learnable}	&68.45&	67.78&	66.88	&65.94 &67.26\\
TE \cite{dong2021exploring}	&58.29	&57.79	&57.18	&56.64 &57.48\\
SAT\cite{madry2017towards}	&61.46	&61.09	&60.47	&59.73 &60.68\\
    \bottomrule
  \end{tabular}
        }
    \end{subtable}  
\end{table}



\begin{table}[H]
    \caption{Results of Classification of Colon Pathology Images (PathMNIST) and COVID-19 Detection from CT Images. The metric is classification accuracy (\%), and the largest value in each column is bold. AutoAttack and white-noise attack are performed using two norm settings: L-inf and L2 norm to measure noise level. SAT \cite{madry2017towards} and TE \cite{dong2021exploring} failed to converge on the COVID-19 datset.}
    \label{MedicalClassification}
    \begin{subtable}{.49\linewidth}
      \centering
        \caption{AutoAttack on PathMNIST (L-inf)}
        \label{PathMNIST1AA}
\resizebox{0.95\columnwidth}{!}{
  \begin{tabular}{|l|lllll|}
    \toprule
   Noise level & 0&2/255&4/255&8/255&Avg.\\
    \midrule
STD	&90.37	&0	&0	&0	&22.59\\
    \midrule
IMA (Ours)	&\textbf{83.07}	&\textbf{65.98}	&47.64	&31.23	&\textbf{56.98}\\
MMA \cite{ding2019mma}	&76.49	&65.90	&45.40	&39.19	&56.74\\
GAIRAT\cite{zhang2020geometry}	&54.67	&44.63	&35.33	&21.50	&39.03\\
FAT\cite{zhang2020attacks}	&69.10	&59.59	&\textbf{52.28}	&39.55	&55.13\\
TRADES\cite{zhang2019theoretically}	&56.08	&52.03	&48.63	&42.92	&49.92\\
IAAT\cite{balaji2019instance}	&80.19	&65.96	&46.86	&33.07	&56.52\\
LBGAT\cite{cui2021learnable}	&58.83	&53.70	&50.51	&44.05	&51.77\\
TE \cite{dong2021exploring}	&49.63	&49.45	&49.35	&\textbf{46.05}	&48.62\\
SAT\cite{madry2017towards}	&54.24	&50.77	&47.38	&39.05	&47.86\\
    \bottomrule
  \end{tabular}
  }
    \end{subtable}%
    \begin{subtable}{.49\linewidth}
      \centering
        \caption{White noise on PathMNIST (L-inf)}
        \label{PathMNIST2W}
        \resizebox{.95\columnwidth}{!}{
  \begin{tabular}{|l|lllll|}
    \toprule
   Noise level & 0& 8/255& 16/255& 32/255&Avg.\\
    \midrule
STD	&90.37	&65.32	&21.36	&4.12	&45.29\\
    \midrule
IMA (Ours)	&\textbf{83.07}	&\textbf{81.15}	&\textbf{77.95}	&65.32	&\textbf{76.87}\\
MMA \cite{ding2019mma}	&76.49	&76.36	&76.21	&\textbf{75.84}	&76.22\\
GAIRAT\cite{zhang2020geometry}	&54.67	&50.58	&46.67	&40.90	&48.21\\
FAT\cite{zhang2020attacks}	&69.10	&65.80	&62.50	&55.57	&63.24\\
TRADES\cite{zhang2019theoretically}	&56.08	&54.94	&53.42	&49.15	&53.40\\
IAAT\cite{balaji2019instance}	&80.19	&77.57	&74.37	&65.20	&74.33\\
LBGAT\cite{cui2021learnable}	&58.83	&56.86	&54.72	&45.96	&54.09\\
TE \cite{dong2021exploring}	&49.63	&49.55	&46.25	&44.09	&47.38\\
SAT\cite{madry2017towards}	&54.24	&52.67	&48.17	&42.16	&49.31\\
    \bottomrule
  \end{tabular}
        }
    \end{subtable}

    \begin{subtable}{.49\linewidth}
      \centering
        \caption{AutoAttack on PathMNIST (L2)}
        \label{PathMNIST1}
\resizebox{0.95\columnwidth}{!}{
  \begin{tabular}{|l|lllll|}
    \toprule
   Noise level & 0&0.3&0.6&0.9&Avg.\\
    \midrule
STD	&90.37&	0&	0	&0&	22.59\\
    \midrule
IMA (Ours)	&\textbf{83.07}&	65.40&	46.66	&36.03&	\textbf{57.79}\\
MMA \cite{ding2019mma}	&76.49&	56.32&	40.12&	30.75&	50.92\\
GAIRAT\cite{zhang2020geometry}	&54.67&	30.66&	14.22&	4.55&	26.03\\
FAT\cite{zhang2020attacks}	&69.10&	47.14&	27.81&	6.99&	37.76\\
TRADES\cite{zhang2019theoretically}	&56.08&	51.88&	48.55&	45.52&	50.51\\
IAAT\cite{balaji2019instance}	&80.19&	\textbf{67.69}&	48.86&	33.03&	57.44\\
LBGAT\cite{cui2021learnable}	&58.83&	53.59&	50.37&	47.01&	52.45\\
TE \cite{dong2021exploring}	&49.63&	49.45&	\textbf{49.35}&	\textbf{49.12}&	49.39\\
SAT\cite{madry2017towards}	&54.24&	50.64&	47.20&	44.67&	49.19\\
    \bottomrule
  \end{tabular}
  }
    \end{subtable}%
    \begin{subtable}{.49\linewidth}
      \centering
        \caption{White noise on PathMNIST (L2)}
        \label{PathMNIST2}
        \resizebox{.95\columnwidth}{!}{
  \begin{tabular}{|l|lllll|}
    \toprule
   Noise level & 0&1&2&3&Avg.\\
    \midrule
STD	&90.37&	56.72&	17.52&	3.74&	42.09\\
    \midrule
IMA (Ours)	&\textbf{83.07}&	\textbf{80.72}&	\textbf{76.51}&	\textbf{69.81}&	\textbf{77.53}\\
MMA \cite{ding2019mma}	&76.49&	71.50&	64.65&	57.59&	67.56\\
GAIRAT\cite{zhang2020geometry}	&54.67&	50.00&	45.87&	42.47&	48.25\\
FAT\cite{zhang2020attacks}	&69.10&	65.22&	61.77&	57.55&	63.41\\
TRADES\cite{zhang2019theoretically}	&56.08&	54.83&	53.05&	51.11&	53.77\\
IAAT\cite{balaji2019instance}	&80.19&	77.57	&74.37&	65.20&	74.33\\
LBGAT\cite{cui2021learnable}	&58.83	&56.54	&54.10&	49.12&	54.65\\
TE \cite{dong2021exploring}	&49.63	&49.52&	45.65&	44.37&	47.29\\
SAT\cite{madry2017towards}	&54.24&	52.49&	47.17&	43.94&	49.46\\
    \bottomrule
  \end{tabular}
        }
    \end{subtable} 

    \begin{subtable}{.49\linewidth}
      \centering
        \caption{AutoAttack on COVID-19 (L-inf)}
        \label{COVID1AA}
\resizebox{0.95\columnwidth}{!}{
  \begin{tabular}{|l|lllll|}
    \toprule
   Noise level & 0&2/255&4/255&8/255&Avg.\\
    \midrule
STD	&97.00	&12.50	&0	&0	&27.38\\
    \midrule
IMA (Ours)	&\textbf{95.25}	&93.75	&92.50&	\textbf{90.00}&	\textbf{92.88}\\
MMA \cite{ding2019mma}	&94.25	&\textbf{94.00}	&\textbf{92.75}&	89.50	&92.63\\
GAIRAT\cite{zhang2020geometry}	&92.00&	90.75	&90.00&	88.50	&90.31\\
FAT\cite{zhang2020attacks}	&88.00	&87.25	&86.25	&81.50	&85.75\\
TRADES\cite{zhang2019theoretically}	&92.75	&91.75	&90.75&	\textbf{90.00}&	91.31\\
IAAT\cite{balaji2019instance}	&\textbf{95.25}	&92.50	&85.00	&64.50	&84.31\\
LBGAT\cite{cui2021learnable}	&78.50	&75.75	&73.00	&64.00	&72.81\\
    \bottomrule
  \end{tabular}}
    \end{subtable} 
    \begin{subtable}{.49\linewidth}
      \centering
        \caption{White noise on COVID-19 (L-inf)}
        \label{COVID2}
        \resizebox{.95\columnwidth}{!}{
  \begin{tabular}{|l|lllll|}
    \toprule
   Noise level & 0& 8/255& 16/255& 32/255&Avg.\\
    \midrule
STD	&97.00&	93.00&	89.75	&85.00&	91.19\\
    \midrule
IMA (Ours)	&\textbf{95.25}	&\textbf{95.25}	& \textbf{94.75}&	\textbf{94.00}	&\textbf{94.81}\\
MMA \cite{ding2019mma}	&94.25&	94.25	&94.25	&\textbf{94.00}	&94.19\\
GAIRAT\cite{zhang2020geometry}	&92.00	&91.50	&90.50	&89.50	&90.88\\
FAT\cite{zhang2020attacks}	&88.00	&87.50	&87.50	&87.50	&87.63\\
TRADES\cite{zhang2019theoretically}	&92.75	&91.75	&91.00	&90.50	&91.50\\
IAAT\cite{balaji2019instance}	&\textbf{95.25}	&94.50	&94.50	&\textbf{94.00}	&94.56\\
LBGAT\cite{cui2021learnable}	&78.54	&77.75	&77.75	&77.00	&77.76\\
    \bottomrule
  \end{tabular}
        }
    \end{subtable}  
    \begin{subtable}{.49\linewidth}
      \centering
        \caption{AutoAttack on COVID-19 (L2)}
        \label{COVID1AA2}
\resizebox{0.95\columnwidth}{!}{
  \begin{tabular}{|l|lllll|}
    \toprule
   Noise level &0& 1 &2 &3 &Avg.\\
    \midrule
STD	&97.00&	9.25&	0	&0	&26.56\\
    \midrule
IMA (Ours)	&\textbf{95.25}	&93.50&	\textbf{91.75}&	\textbf{91.00}&	\textbf{92.88}\\
MMA \cite{ding2019mma}	&94.25&	\textbf{93.75}&	\textbf{91.75}	&90.00&	92.44\\
GAIRAT\cite{zhang2020geometry}	&92.00	&84.50&	73.00&	62.25&	77.94\\
FAT\cite{zhang2020attacks}	&88.00	&68.00&	51.50&	31.75&	59.81\\
TRADES\cite{zhang2019theoretically}	&92.75&	91.25	&90.50	&89.50&	91.00\\
IAAT\cite{balaji2019instance}	&\textbf{95.25}&	91.50&	82.25&	70.25&	84.81\\
LBGAT\cite{cui2021learnable}	&78.50&	76.00	&72.75&	68.25&	73.88\\
    \bottomrule
  \end{tabular}
  }
    \end{subtable} 
    \begin{subtable}{.49\linewidth}
      \centering
        \caption{White noise on COVID-19 (L2)}
        \label{COVIDW2}
        \resizebox{.95\columnwidth}{!}{
  \begin{tabular}{|l|lllll|}
    \toprule
   Noise level & 0& 9&18&27&Avg.\\
    \midrule
STD	&97.00&	95.00&	92.75&	88.00&	93.19\\
    \midrule
IMA (Ours)	&\textbf{95.25}&	\textbf{95.25}&	\textbf{95.00}&	\textbf{94.75}&	\textbf{95.06}\\
MMA \cite{ding2019mma}	&94.25&	94.25&	94.25&	94.25&	94.25\\
GAIRAT\cite{zhang2020geometry}	&92.00&	91.25&	90.50&	90.00&	90.94\\
FAT\cite{zhang2020attacks}	&88.00&	87.75&	87.50&	87.50&	87.69\\
TRADES\cite{zhang2019theoretically}	&92.75&	92.75&	91.75&	90.75&	92.00\\
IAAT\cite{balaji2019instance}	&\textbf{95.25}&	95.00&	94.50&	94.50&	94.81\\
LBGAT\cite{cui2021learnable}	&78.50&	78.00&	77.75&	77.50&	77.94\\
    \bottomrule
  \end{tabular}
        }
    \end{subtable}
\end{table} 


\begin{table}[H]
    \caption{Results of the medical image segmentation applications on Heart, Hippocampus, and Prostate datasets. The metric is Dice score (\%), and the largest value in each column is bold. AutoAttack and white-noise attack are performed using two norm settings: L-inf and L2 norm to measure noise level.}
    \label{Segmentation}
    \begin{subtable}{.49\linewidth}
      \centering
        \caption{AutoAttack on Heart (L-inf)}
        \label{Heart1A}
\resizebox{0.95\columnwidth}{!}{
  \begin{tabular}{|l|lllll|}
    \toprule
   Noise level & 0&2/255&4/255&8/255&Avg.\\
    \midrule
STD	&92.46&	69.14&	63.77&	50.66&	69.01\\
    \midrule
IMA (Ours)	&\textbf{91.94}&	71.58&	\textbf{69.50}&	\textbf{65.01}&	\textbf{74.51}\\
TRADES\cite{zhang2019theoretically}	&91.21&	\textbf{73.17}&	66.33&	52.38&	70.77\\
TE\cite{dong2021exploring} &66.51&	65.61&	65.59&	64.82&	65.63\\
SAT\cite{madry2017towards}	&79.34&	69.14&	66.79&	62.54&	69.45\\
    \bottomrule
  \end{tabular}
  }
    \end{subtable}   
    \begin{subtable}{.49\linewidth}
      \centering
        \caption{White noise on Heart (L-inf)}
        \label{Heart1W}
        \resizebox{.95\columnwidth}{!}{
  \begin{tabular}{|l|lllll|}
    \toprule
   Noise level & 0& 8/255& 16/255&32/255&Avg.\\
    \midrule
STD	&92.46&	73.27&	71.00&	70.79&	76.88\\
    \midrule
IMA (Ours)	&\textbf{91.94}&	73.64&	\textbf{73.54}&	\textbf{73.48}&	\textbf{78.15}\\
TRADES\cite{zhang2019theoretically}	&91.21&	\textbf{75.81}&	73.23&	72.23&	78.12\\
TE\cite{dong2021exploring} &66.51&	65.62&	64.73&	64.42&	65.32\\
SAT\cite{madry2017towards}	&79.34&	70.99&	70.13&	68.93&	72.35\\
 \bottomrule
  \end{tabular}
        }
    \end{subtable}

    \begin{subtable}{.49\linewidth}
      \centering
        \caption{AutoAttack on Heart (L2)}
        \label{Heart2A}
\resizebox{0.95\columnwidth}{!}{
  \begin{tabular}{|l|lllll|}
    \toprule
   Noise level & 0	&10	&20	&30&Avg.\\
    \midrule
STD	&92.46&	69.85&	64.48&	60.73&	71.88\\
    \midrule
IMA (Ours)	&\textbf{91.94}&	\textbf{71.36}&	\textbf{68.08}&	64.90&	\textbf{74.07}\\
TRADES\cite{zhang2019theoretically}	&91.21&	70.06&	67.79&	64.31&	73.34\\
TE\cite{dong2021exploring} &66.51&	65.74&	64.89&	\textbf{64.93}&	65.52\\
SAT\cite{madry2017towards}	&79.34&	68.70&	66.32&	64.31&	69.67\\
    \bottomrule
  \end{tabular}
  }
    \end{subtable}   
    \begin{subtable}{.49\linewidth}
      \centering
        \caption{White noise on Heart (L2)}
        \label{Heart2W}
        \resizebox{.95\columnwidth}{!}{
  \begin{tabular}{|l|lllll|}
    \toprule
   Noise level & 0	&10	&20	&30&Avg.\\
    \midrule
STD	&92.46&	73.08&	67.60&	65.60&	74.69\\
    \midrule
IMA (Ours)	&\textbf{91.94}&	73.60&	\textbf{73.29}&	\textbf{73.02}&	\textbf{77.96}\\
TRADES\cite{zhang2019theoretically}	&91.21&	\textbf{75.52}&	71.45&	67.35&	76.38\\
TE\cite{dong2021exploring} &66.51&	64.65&	64.59&	63.82&	64.89\\
SAT\cite{madry2017towards}	&79.34&	70.11&	69.23&	68.29&	71.74\\
 \bottomrule
  \end{tabular}
        }
    \end{subtable}  

    \begin{subtable}{.49\linewidth}
      \centering
        \caption{AutoAttack on Hippocampus (L-inf)}
        \label{Hippocampus1A}
\resizebox{0.95\columnwidth}{!}{
  \begin{tabular}{|l|lllll|}
    \toprule
   Noise level & 0	&2/255	&4/255	&8/255&Avg.\\
    \midrule
STD	&86.62&	71.24&	68.74&	60.97&	71.89\\
    \midrule
IMA (Ours)	&\textbf{85.90}&	\textbf{79.25}&	\textbf{78.86}&	\textbf{78.07}&	\textbf{80.52}\\
TRADES\cite{zhang2019theoretically}	&83.61&	73.39&	71.53&	67.05&	73.90\\
TE\cite{dong2021exploring} &81.67&	78.46&	78.31&	77.86&	79.08\\
    \bottomrule
  \end{tabular}
  }
    \end{subtable}   
    \begin{subtable}{.49\linewidth}
      \centering
        \caption{White noise on Hippocampus (L-inf)}
        \label{Hippocampus1W}
        \resizebox{.95\columnwidth}{!}{
  \begin{tabular}{|l|lllll|}
    \toprule
   Noise level & 0& 8/255& 16/255&32/255&Avg.\\
    \midrule
STD	&86.62&	74.10&	73.97&	73.50&	77.05\\
    \midrule
IMA (Ours)	&\textbf{85.90}&	\textbf{80.10}&	\textbf{79.99}&	\textbf{79.96}&	\textbf{81.49}\\
TRADES\cite{zhang2019theoretically}	&83.61&	76.09&	76.07&	75.58&	77.84\\
TE\cite{dong2021exploring} &81.67&	79.50&	78.20&	78.10&	79.37\\
 \bottomrule
  \end{tabular}
        }
    \end{subtable}  

    \begin{subtable}{.49\linewidth}
      \centering
        \caption{AutoAttack on Hippocampus (L2)}
        \label{Hippocampus2A}
\resizebox{0.95\columnwidth}{!}{
  \begin{tabular}{|l|lllll|}
    \toprule
   Noise level & 0&5&15&25&Avg.\\
    \midrule
STD	&86.62&	66.65&	28.24&	0.10&	45.40\\
    \midrule
IMA (Ours)	&\textbf{85.90}&	\textbf{78.01}&	\textbf{70.49}&	\textbf{30.32}&	\textbf{66.18}\\
TRADES\cite{zhang2019theoretically}	&83.61&	69.65&	41.69&	0.27&	48.81\\
TE\cite{dong2021exploring} &81.67&	77.77&	68.94&	17.02&	61.35\\
    \bottomrule
  \end{tabular}
  }
    \end{subtable}   
    \begin{subtable}{.49\linewidth}
      \centering
        \caption{White noise on Hippocampus (L2)}
        \label{Hippocampus2W}
        \resizebox{.95\columnwidth}{!}{
  \begin{tabular}{|l|lllll|}
    \toprule
   Noise level & 0&5&15&25&Avg.\\
    \midrule
STD	&86.62&	72.61&	69.70&	64.58&	73.38\\
    \midrule
IMA (Ours)	&\textbf{85.90}&	\textbf{82.19}&	\textbf{81.66}&	\textbf{80.44}&	\textbf{82.55}\\
TRADES\cite{zhang2019theoretically}	&83.61&	75.33&	72.90&	67.62&	74.87\\
TE\cite{dong2021exploring} &81.67&	80.69&	80.59&	78.74&	80.42\\
 \bottomrule
  \end{tabular}
        }
    \end{subtable}  

    \begin{subtable}{.49\linewidth}
      \centering
        \caption{AutoAttack on Prostate (L-inf)}
        \label{Prostate1A}
\resizebox{0.95\columnwidth}{!}{
  \begin{tabular}{|l|lllll|}
    \toprule
   Noise level & 0&2/255&4/255&8/255&Avg.\\
    \midrule
STD	&81.02&	67.50&	63.14&	53.74&	66.35\\
    \midrule
IMA (Ours)	&\textbf{87.85}&	\textbf{84.91}&	\textbf{84.46}&	\textbf{83.45}&	\textbf{85.17}\\
TRADES\cite{zhang2019theoretically}	&80.21&	72.75&	69.48&	61.82&	71.07\\
TE\cite{dong2021exploring} &78.73&	69.82&	69.06&	67.39&	71.25\\
SAT\cite{madry2017towards}	&69.55&	67.07&	66.26&	64.59&	66.87\\
    \bottomrule
  \end{tabular}
  }
    \end{subtable}   
    \begin{subtable}{.49\linewidth}
      \centering
        \caption{White noise on Prostate (L-inf)}
        \label{Prostate1W}
        \resizebox{.95\columnwidth}{!}{
  \begin{tabular}{|l|lllll|}
    \toprule
   Noise level & 0& 8/255& 16/255&32/255&Avg.\\
    \midrule
STD	&81.02&	72.37&	71.88&	71.55&	74.21\\
    \midrule
IMA (Ours)	&\textbf{87.85}&	\textbf{85.85}&	\textbf{85.76}&	\textbf{85.69}&	\textbf{86.29}\\
TRADES\cite{zhang2019theoretically}	&80.21&	76.40&	76.08&	75.89&	77.15\\
TE\cite{dong2021exploring} &78.73&	77.03&	77.02&	76.94&	77.43\\
SAT\cite{madry2017towards}	&69.55&	68.81&	68.51&	68.48&	68.84\\
 \bottomrule
  \end{tabular}
        }
    \end{subtable}  

    \begin{subtable}{.49\linewidth}
      \centering
        \caption{AutoAttack on Prostate (L2)}
        \label{Prostate2A}
\resizebox{0.95\columnwidth}{!}{
  \begin{tabular}{|l|lllll|}
    \toprule
   Noise level & 0	&10	&20	&30&Avg.\\
    \midrule
STD	&81.02&	69.16&	67.86&	65.66&	70.93\\
    \midrule
IMA (Ours)	&\textbf{87.85}&	\textbf{85.07}&	\textbf{84.71}&	\textbf{84.37}&	\textbf{85.50}\\
TRADES\cite{zhang2019theoretically}	&80.21&	74.16&	72.72&	71.97&	74.77\\
TE\cite{dong2021exploring} &78.73&	70.10&	69.72&	69.29&	71.96\\
SAT\cite{madry2017towards}	&69.55&	67.24&	66.85&	66.20&	67.46\\
    \bottomrule
  \end{tabular}
  }
    \end{subtable}   
    \begin{subtable}{.49\linewidth}
      \centering
        \caption{White noise on Prostate (L2)}
        \label{Prostate2W}
        \resizebox{.95\columnwidth}{!}{
  \begin{tabular}{|l|lllll|}
    \toprule
   Noise level & 0	&10	&20	&30&Avg.\\
    \midrule
STD	&81.02&	72.60&	72.00&	71.37&	74.25\\
    \midrule
IMA (Ours)	&\textbf{87.85}&	\textbf{85.88}&	\textbf{85.76}&	\textbf{85.27}&	\textbf{86.19}\\
TRADES\cite{zhang2019theoretically}	&80.21&	76.27&	76.16&	75.92&	77.14\\
TE\cite{dong2021exploring}          &78.73&	77.08&	77.08&	76.91&	77.45\\
SAT\cite{madry2017towards}	        &69.55&	68.71&	68.61&	68.52&	68.85\\
 \bottomrule
  \end{tabular}
        }
    \end{subtable}  
\end{table}

\section{Discussion}
\label{4.2}

From Tables \ref{CIFARClassification} - \ref{MedicalClassification}, we can make these conclusions: (1) IMA outperforms the other eight representative defense methods on the image classification applications. (2) IMA has the least accuracy degradation on clean data.

From Table \ref{Segmentation}, we can make these conclusions: (1) IMA outperforms other methods in general for image segmentation applications. (2) IMA always has the least accuracy degradation on clean data. This demonstrates that IMA can generate nearly-optimal adversarial training samples such that the model's standard accuracy can be preserved as much as possible (even improved in the Prostate application), which is a unique property of IMA and desirable for medical applications.

In addition, from all the results in Table \ref{CIFARClassification}, Table 
\ref{MedicalClassification}, and Table \ref{Segmentation}, a DNN model that is more robust to adversarial noises is also more robust to random noises. This conclusion is also supported by the observation in \cite{fawzi2016robustness}.

Although IMA has the best performance on both standard accuracy and adversarial robustness, it still shows reduction in standard accuracy for the applications (except the Prostate application). To further improve the performance, better algorithms will be needed to more accurately estimate sample margins, which warrants a future study. 

We would like to discuss the differences and similarities between adversarial training and generative adversarial learning, although both approaches belong to the same category of adversarial learning. Generative adversarial learning trains a discriminator and a generator together such that the generator is able to generate high-quality samples that can fool the discriminator, and this approach can improve the standard accuracy of DNNs by generating high-quality training samples \cite{jeong2022robust, qiu2022improved, salvi2022dermocc, hazra2022enhancing, pham2022generating}. As a comparison, adversarial training methods generate adversarial training samples to improve the adversarial/noise robustness of DNNs. Our method can generate high-quality adversarial training samples to preserve standard accuracy while improving adversarial robustness.

\section{Conclusion}

In this study, we design IMA, a novel adversarial training method focusing not only on adversarial robustness but also on standard accuracy. We evaluate IMA on six publicly available image datasets under AutoAttack and white-noise attack, and the results show IMA outperforms the other defense methods for classification and segmentation. The result of the prostate image segmentation application shows that both adversarial robustness and standard accuracy are improved by IMA. To our knowledge, for the first time, we demonstrate that it is possible to avoid the trade-off between standard accuracy and adversarial robustness for medical image segmentation, which was thought to be mission impossible. Our work will facilitate the development of robust applications in the medical field.

\section{Acknowledgement}
This work was supported in part by the NIH grant R01HL158829.

\appendix

\section {Optimal Adversarial Training Samples}

\begin{figure}[ht]
\centering
\begin{subfigure}[b]{0.3\textwidth}
\includegraphics[width=0.98\linewidth]{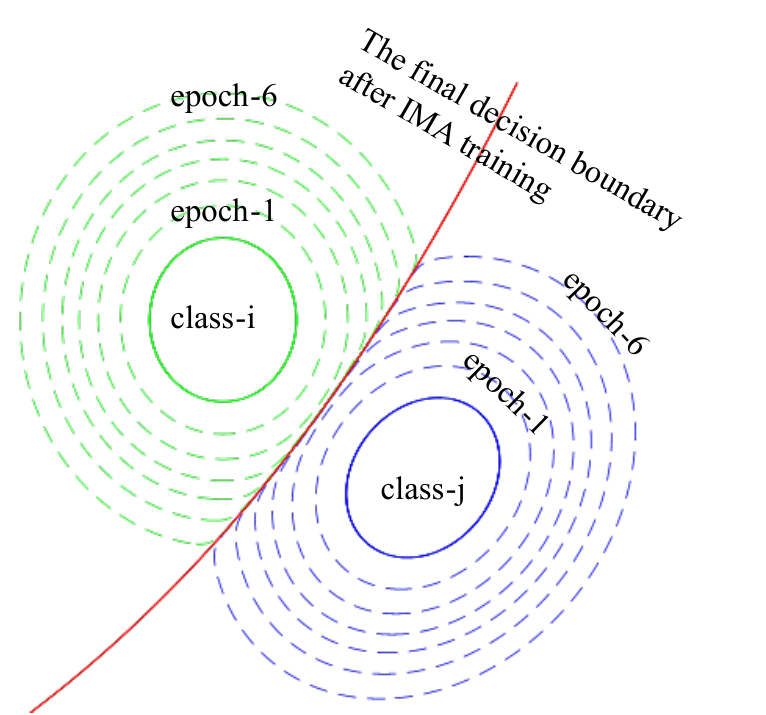}
\caption{}
\end{subfigure}
\begin{subfigure}[b]{0.3\textwidth}
\includegraphics[width=0.98\linewidth]{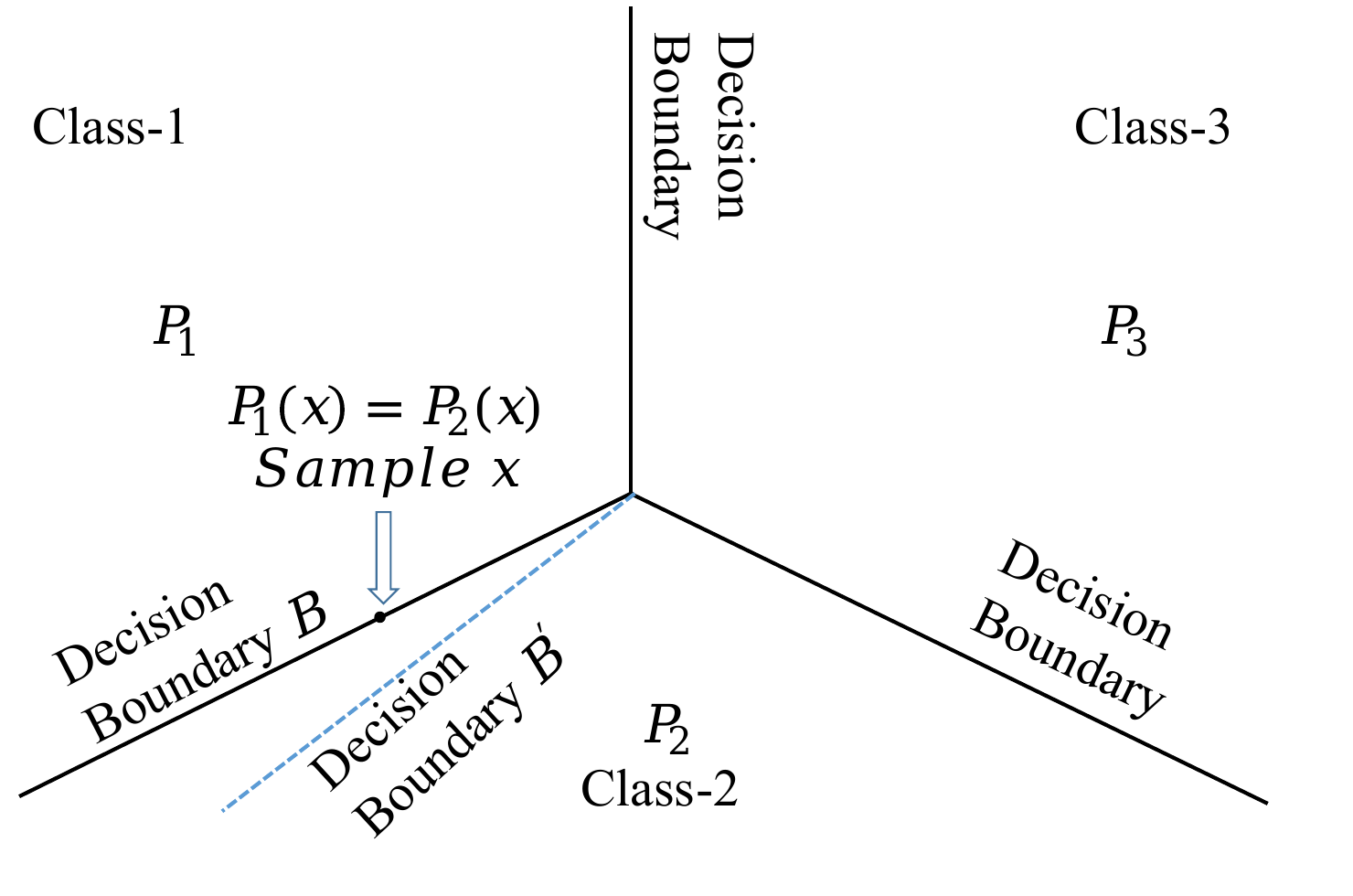}
\caption{}
\end{subfigure}
\caption{ (a) The mechanism of gradual margin expansion is illustrated from the perspective of the samples in two classes. During training, the sample-wise estimated margins $\mathcal{E}$ are expanded epoch by epoch until the algorithm converges. (b) The Equilibrium State is shown in the case of three classes. These are intuitive explanations, and our method does not assume linear separability between classes.}
\label{fig4}
\end{figure}

Here, we show that if the adversarial training samples are placed near the decision boundaries, then the adversarial training loss will be minimized.


To simplify the discussion, we assume there are three classes and three decision boundaries between classes (Fig. \ref{fig4}). The softmax output of the DNN model $f$ has three components: $p_1$, $p_2$ and $p_3$ corresponding to the three classes. If a data point $\tilde{x}$ is about to cross the decision boundary $B_{ij}$ between class-$i$ $(c_i)$ and class-$j$ $(c_j)$, then $p_i(\tilde{x})=p_j(\tilde{x})$. The mathematical expectation of the cross-entropy loss of the generated adversarial training samples (i.e., $L_1$ in Algorithm \ref{alg1}, when $x$ is correctly classified) is:

\begin{scriptsize}
\begin{equation}
\scriptsize
\label{eq3}
E=\mathbf{E}_{\tilde{x}\in c_1}\left(-log\left(p_1\left(\tilde{x}\right)\right)\right) 
 +\mathbf{E}_{\tilde{x}\in c_2}\left(-log\left(p_2\left(\tilde{x}\right)\right)\right) 
 +\mathbf{E}_{\tilde{x}\in c_3}\left(-log\left(p_3\left(\tilde{x}\right)\right)\right)
\end{equation}
\end{scriptsize}

In the above equation, for simplicity, we also use $\tilde{x}$ to represent a random variable/sample. The IMA method pushes the adversarial training samples toward the decision boundaries. For each sample $\tilde{x} \in c_i$, it is on the class-$i$ side of a decision boundary. We use $\tilde{x} \in c_i \cap B_{ij}$ to denote $\tilde{x} \in c_i$ and $\tilde{x}$ is on the class-$i$ side of the decision boundary $B_{ij}$. Thus, we can obtain:

\begin{scriptsize}
\begin{equation}
\scriptsize
\mathbf{E}_{\tilde{x}\in c_1}\left(-log\left(p_1\left(\tilde{x}\right)\right)\right)=\mathbf{E}_{\tilde{x}\in c_1 \cap B_{12}}\left(-log\left(p_1\left(\tilde{x}\right)\right)\right) 
+\mathbf{E}_{\tilde{x}\in c_1 \cap B_{13}}\left(-log\left(p_1\left(\tilde{x}\right)\right)\right) 
\end{equation}
\end{scriptsize}
\begin{scriptsize}
\begin{equation}
\scriptsize
\mathbf{E}_{\tilde{x}\in c_2}\left(-log\left(p_2\left(\tilde{x}\right)\right)\right)=\mathbf{E}_{\tilde{x}\in c_2 \cap B_{12}}\left(-log\left(p_2\left(\tilde{x}\right)\right)\right) 
+\mathbf{E}_{\tilde{x}\in c_2 \cap B_{23}}\left(-log\left(p_2\left(\tilde{x}\right)\right)\right) 
\end{equation}
\end{scriptsize}
\begin{scriptsize}
\begin{equation}
\scriptsize
\mathbf{E}_{\tilde{x}\in c_3}\left(-log\left(p_3\left(\tilde{x}\right)\right)\right)=\mathbf{E}_{\tilde{x}\in c_3 \cap B_{13}}\left(-log\left(p_3\left(\tilde{x}\right)\right)\right)
+\mathbf{E}_{\tilde{x}\in c_3 \cap B_{23}}\left(-log\left(p_3\left(\tilde{x}\right)\right)\right)
\end{equation}
\end{scriptsize}

If the generated adversarial training samples (random variables) \{$\tilde{x}\ \vert\ \tilde{x}\in c_i$\} and \{$\tilde{x}\ \vert\ \tilde{x}\in c_j$\} have the same spatial distribution on the decision boundary $B_{ij}$ between the two classes, then:

\begin{scriptsize}
\begin{equation}
\scriptsize
\begin{split}
&\mathbf{E}_{\tilde{x}\in c_i \cap B_{ij}} (-log(p_i(\tilde{x}))) + \mathbf{E}_{\tilde{x}\in c_j \cap B_{ij}} (-log(p_j(\tilde{x})))\\
&=\mathbf{E}_{\tilde{x}\in\ B_{ij}}\left(-log\left(p_i\left(\tilde{x}\right)\right)-log\left(p_j\left(\tilde{x}\right)\right)\right) \\
&=\mathbf{E}_{\tilde{x}\in\ B_{ij}}\left(-log\left(p_i\left(\tilde{x}\right)p_j\left(\tilde{x}\right)\right)\right) \\
&\geq \mathbf{E}_{\tilde{x}\in\ B_{ij}} (-log (\frac{p_i\left(\tilde{x}\right)+p_j\left(\tilde{x}\right)}{2})^2)
\end{split}
\end{equation}
\end{scriptsize}

As a result, $E$ reaches the minimum when $p_i\left(\tilde{x}\right)=p_j\left(\tilde{x}\right)$. This indicates that the optimal adversarial training samples should be close to the decision boundary of the classifier, which will lead to an equilibrium state.

\section{Selection of $\Delta_\epsilon$ for classification}
\label{A1}

\begin{figure}[H]
  \centering
  \begin{subfigure}{0.3\textwidth}
    \includegraphics[width=0.98\linewidth]{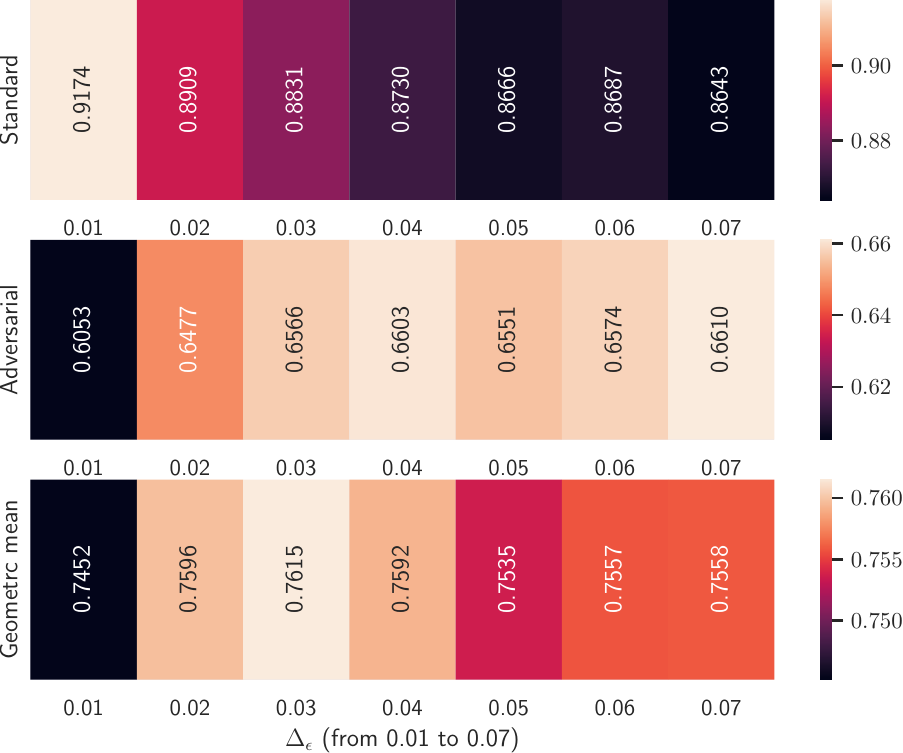}
    \caption{CIFAR10 }
    \end{subfigure}
  \begin{subfigure}{0.3\textwidth}
    \includegraphics[width=0.98\linewidth]{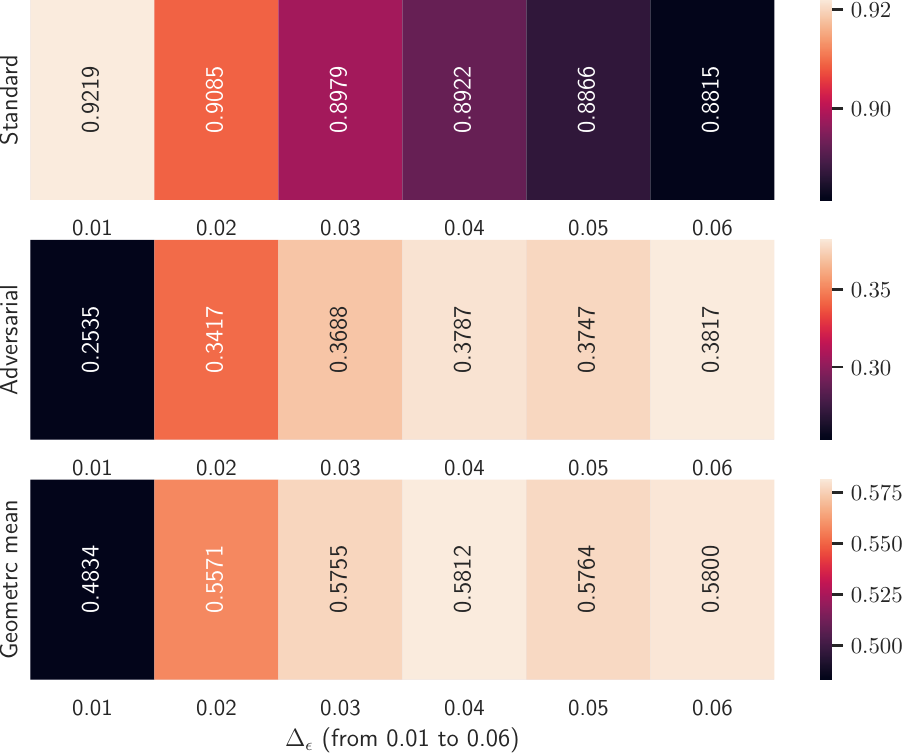}
    \caption{PathMNIST}
    \end{subfigure}
  \begin{subfigure}{0.3\textwidth}
    \includegraphics[width=0.98\linewidth]{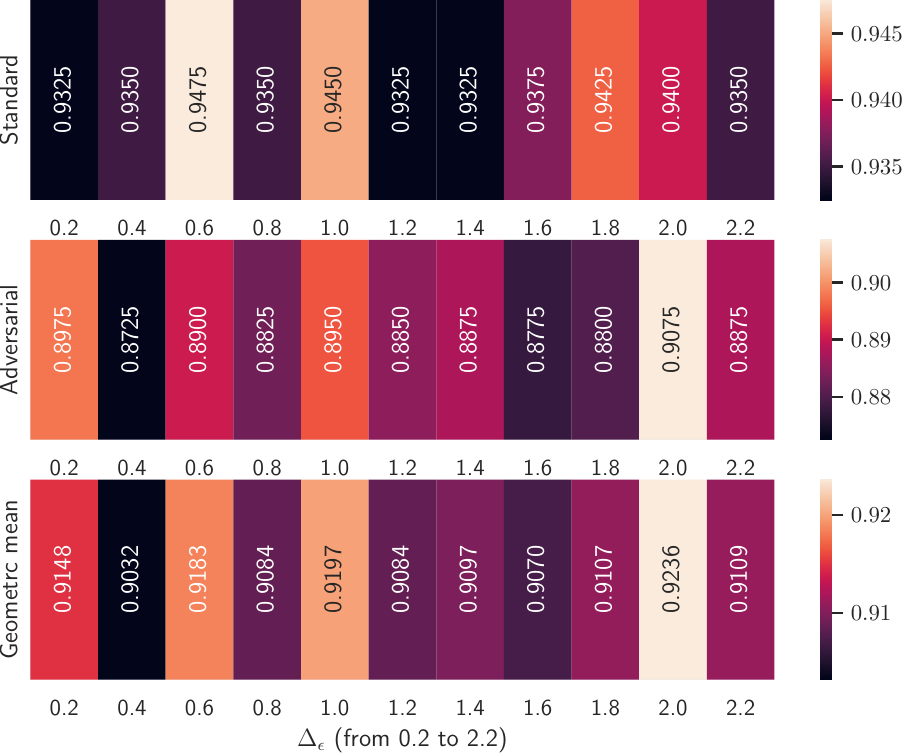}
    \caption{COVID-19}
    \end{subfigure}
  \caption{The effect of step size $\Delta_{\epsilon}$ in IMA (given $\epsilon_{train}=3$). Under different $\Delta_{\epsilon}$: the first row shows the standard accuracy; the second row shows the adversarial accuracy under PGD attack with noise level 0.5; the third row is the geometric mean of the first two rows, which can be used as the metric to select the optimal $\Delta_\epsilon$.}
  \label{hyper1}
\end{figure}

Fig. \ref{hyper1} shows how the step size $\Delta_{\epsilon}$ affects the performance of IMA. As the $\Delta_{\epsilon}$ increases, standard accuracy and adversarial accuracy stay stable, except for too small $\Delta_{\epsilon}$. This is because IMA with too small $\Delta_{\epsilon}$ may need a larger number of training epochs to converge, while the grid research has a fixed number of epochs for each run. The geometric mean of the two accuracies can be used for hyper-parameter selection. In this case, the best $\Delta_{\epsilon}$ is 0.03 for CIFAR10, 0.04 for PathMNIST and 2.0 for COVID-19.

\section{Selection of $\Delta_\epsilon$ for segmentation}
\label{A2}

\begin{figure}[H]
  \centering
  \begin{subfigure}{0.3\textwidth}
    \includegraphics[width=0.98\linewidth]{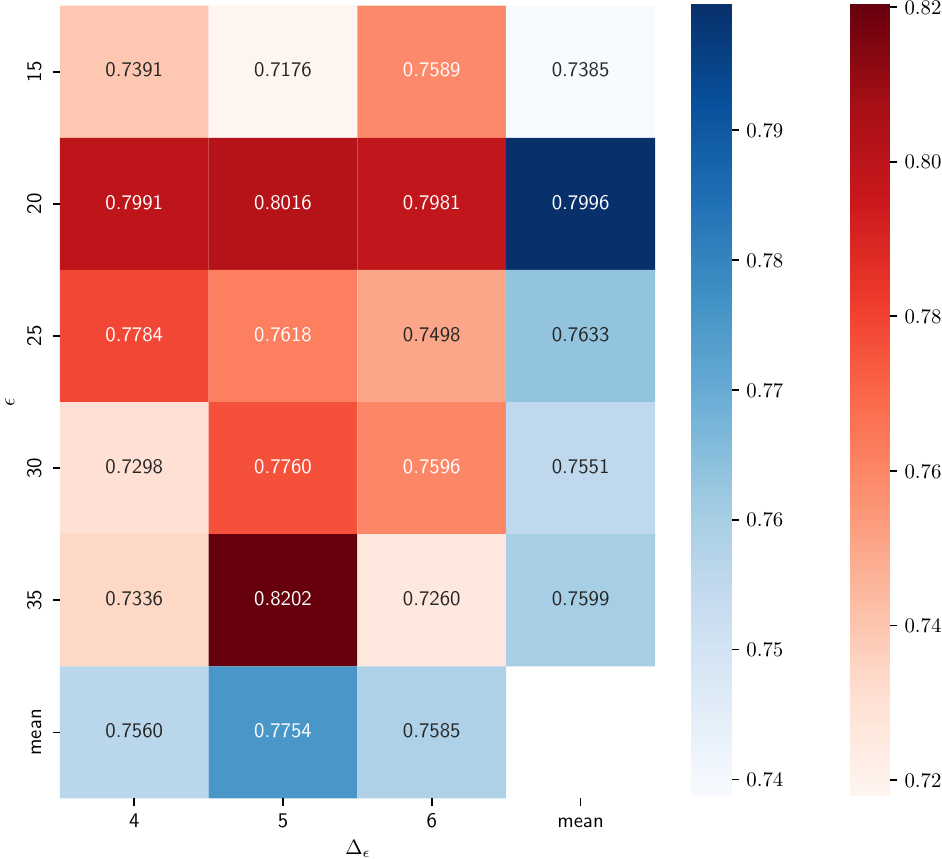}
    \caption{Heart}
    \end{subfigure}
  \begin{subfigure}{0.3\textwidth}
    \includegraphics[width=0.98\linewidth]{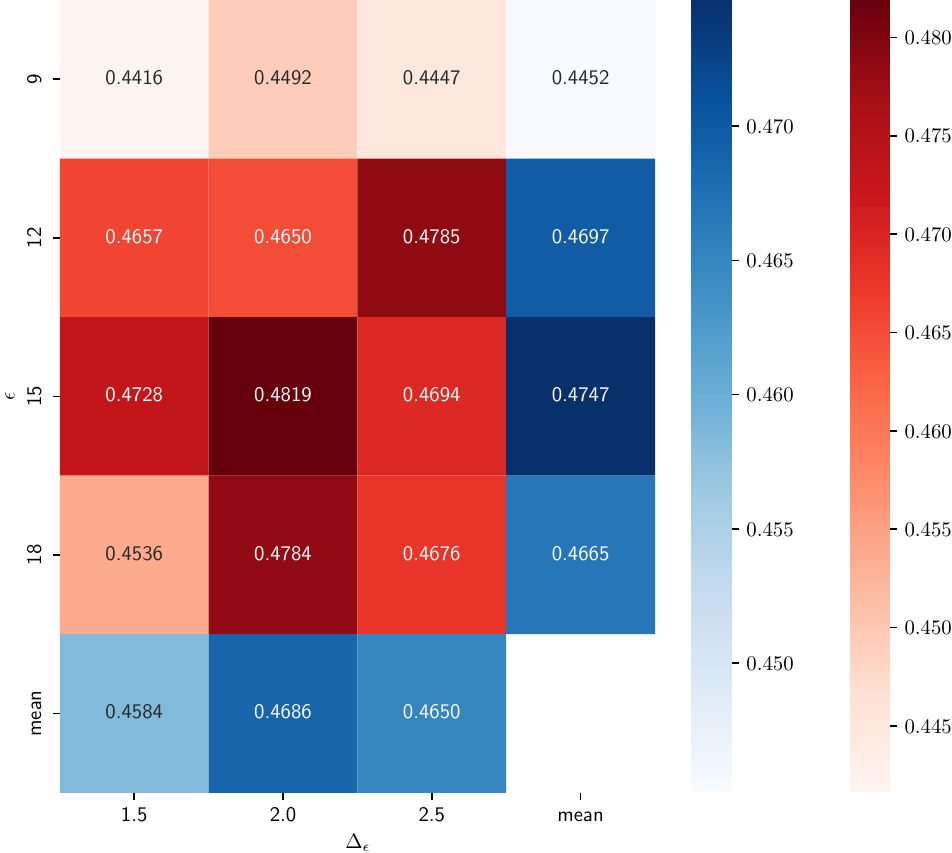}
    \caption{Hippocampus}
    \end{subfigure}
  \begin{subfigure}{0.3\textwidth}
    \includegraphics[width=0.98\linewidth]{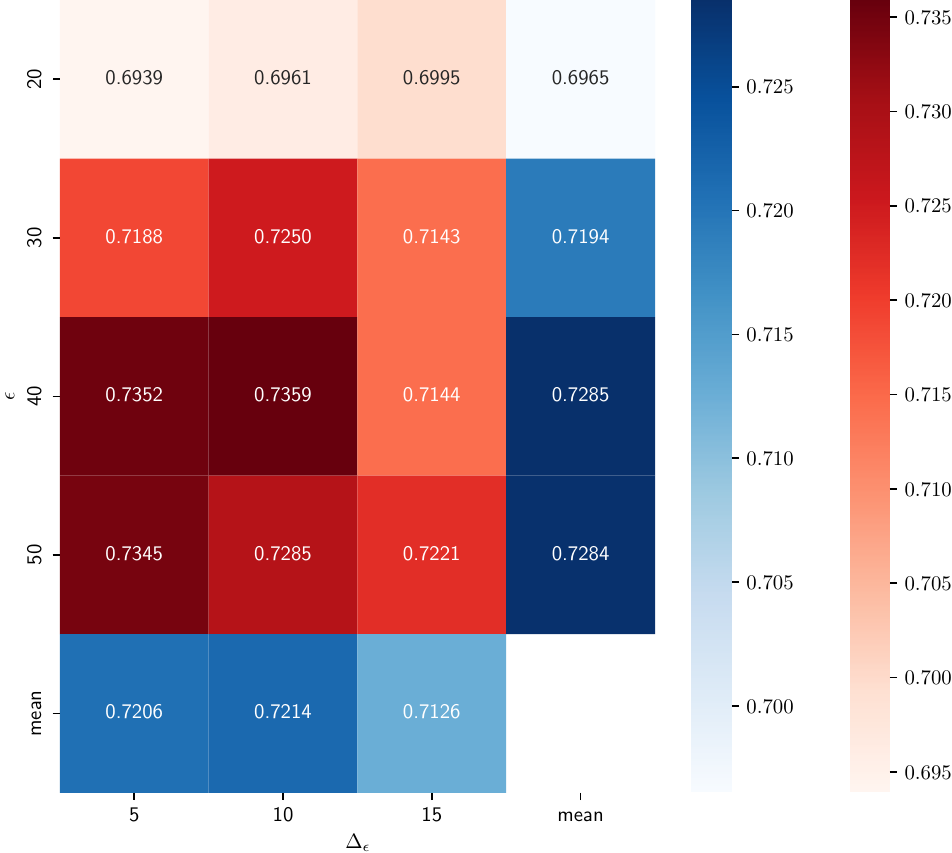}
    \caption{Prostate}
    \end{subfigure}
  \caption{The effect of step size $\Delta_{\epsilon}$ and $\epsilon$ in the IMA. The X-axis is $\Delta_\epsilon$. The Y-axis is $\epsilon$. Each entry shows the geometric mean of the standard accuracy and adversarial accuracy under the corresponding pair of  $\Delta_{\epsilon}$ and $\epsilon$. The optimal $\Delta_{\epsilon}$ leads to the largest value in the table.}
  \label{hyper2}
\end{figure}

Fig. \ref{hyper2} shows the result of the grid search for IMA. The geometric mean of the standard accuracy and adversarial accuracy can be used for hyperparameter selection. The optimal $\Delta_{\epsilon}$ values are 5 for Heart, 2 for Hippocampus, and 10 for Prostate.

\begin{singlespace}
\bibliographystyle{elsarticle-num} 
\bibliography{IMA}

\begin{thebibliography}{10}
\expandafter\ifx\csname url\endcsname\relax
  \def\url#1{\texttt{#1}}\fi
\expandafter\ifx\csname urlprefix\endcsname\relax\def\urlprefix{URL }\fi
\expandafter\ifx\csname href\endcsname\relax
  \def\href#1#2{#2} \def\path#1{#1}\fi

\bibitem{szegedy2014intriguing}
C.~Szegedy, W.~Zaremba, et~al., Intriguing properties of neural networks, in:
  International Conference on Learning Representations, 2014.

\bibitem{goodfellow2014explaining}
I.~Goodfellow, J.~Shlens, et~al., Explaining and harnessing adversarial
  examples, in: International Conference on Learning Representations, 2015.

\bibitem{akhtar2018threat}
N.~Akhtar, A.~Mian, Threat of adversarial attacks on deep learning in computer
  vision: A survey, IEEE Access (2018).

\bibitem{graese2016assessing}
A.~Graese, A.~Rozsa, et~al., Assessing threat of adversarial examples on deep
  neural networks, in: IEEE International Conference on Machine Learning and
  Applications, 2016.

\bibitem{mirjalili2017soft}
V.~Mirjalili, A.~Ross, Soft biometric privacy: Retaining biometric utility of
  face images while perturbing gender, in: IEEE International Joint Conference
  on Biometrics, 2017.

\bibitem{eykholt2018robust}
K.~Eykholt, I.~Evtimov, et~al., Robust physical-world attacks on deep learning
  visual classification, in: Proceedings of the IEEE conference on computer
  vision and pattern recognition, 2018, pp. 1625--1634.

\bibitem{WHO2020Coronavirus}
Coronavirus disease (covid-19) dashboard, \url{https://covid19.who.int/}
  (2020).

\bibitem{ai2020correlation}
T.~Ai, Z.~Yang, et~al., Correlation of chest ct and rt-pcr testing in
  coronavirus disease 2019 (covid-19) in china: a report of 1014 cases,
  Radiology (2020).

\bibitem{shi2020review}
F.~Shi, J.~Wang, et~al., Review of artificial intelligence techniques in
  imaging data acquisition, segmentation and diagnosis for covid-19, IEEE
  Reviews in Biomedical Engineering (2020).

\bibitem{he2016deep}
K.~He, X.~Zhang, et~al., Deep residual learning for image recognition, in:
  Proceedings of the IEEE conference on computer vision and pattern
  recognition, 2016.

\bibitem{fawzi2016robustness}
A.~Fawzi, S.-M. Moosavi-Dezfooli, et~al., Robustness of classifiers: from
  adversarial to random noise, in: Conference on Neural Information Processing
  Systems, 2016.

\bibitem{gilmer2019adversarial}
J.~Gilmer, N.~Ford, et~al., Adversarial examples are a natural consequence of
  test error in noise, in: International Conference on Machine Learning, PMLR,
  2019, pp. 2280--2289.

\bibitem{madry2017towards}
A.~Madry, A.~Makelov, et~al., Towards deep learning models resistant to
  adversarial attacks, in: International Conference on Learning
  Representations, 2018.

\bibitem{kurakin2016adversarial}
A.~Kurakin, I.~Goodfellow, et~al., Adversarial examples in the physical world,
  in: Artificial intelligence safety and security, 2018.

\bibitem{dong2021exploring}
Y.~Dong, K.~Xu, et~al., Exploring memorization in adversarial training,
  International Conference on Learning Representations (2022).

\bibitem{jia2022boosting}
X.~Jia, Y.~Zhang, et~al., Boosting fast adversarial training with learnable
  adversarial initialization, IEEE Transactions on Image Processing 31 (2022)
  4417--4430.

\bibitem{zhang2019theoretically}
H.~Zhang, Y.~Yu, othersl, Theoretically principled trade-off between robustness
  and accuracy, in: International Conference on Machine Learning, 2019.

\bibitem{wang2019improving}
Y.~Wang, D.~Zou, et~al., Improving adversarial robustness requires revisiting
  misclassified examples, in: International Conference on Learning
  Representations, 2019.

\bibitem{wang2019convergence}
Y.~Wang, X.~Ma, et~al., On the convergence and robustness of adversarial
  training, in: International Conference on Machine Learning, 2019.

\bibitem{sitawarin2020improving}
C.~Sitawarin, S.~Chakraborty, D.~Wagner, Sat: Improving adversarial training
  via curriculum-based loss smoothing, in: Proceedings of the 14th ACM Workshop
  on Artificial Intelligence and Security, 2021, pp. 25--36.

\bibitem{cai2018curriculum}
Q.-Z. Cai, C.~Liu, et~al., Curriculum adversarial training, in: International
  Joint Conference on Artificial Intelligence, 2018.

\bibitem{balaji2019instance}
Y.~Balaji, T.~Goldstein, et~al., Instance adaptive adversarial training:
  Improved accuracy tradeoffs in neural nets, preprint arXiv:1910.08051 (2019).

\bibitem{zhang2020attacks}
J.~Zhang, X.~Xu, et~al., Attacks which do not kill training make adversarial
  learning stronger, in: International Conference on Machine Learning, 2020.

\bibitem{cheng2020cat}
M.~Cheng, Q.~Lei, et~al., Cat: Customized adversarial training for improved
  robustness, preprint arXiv:2002.06789 (2020).

\bibitem{ding2019mma}
G.~W. Ding, Y.~Sharma, et~al., Mma training: Direct input space margin
  maximization through adversarial training, in: International Conference on
  Learning Representations, 2020.

\bibitem{zhang2020geometry}
J.~Zhang, J.~Zhu, et~al., Geometry-aware instance-reweighted adversarial
  training, in: International Conference on Learning Representations, 2020.

\bibitem{cui2021learnable}
J.~Cui, S.~Liu, et~al., Learnable boundary guided adversarial training, in:
  Proceedings of the IEEE International Conference on Computer Vision, 2021,
  pp. 15721--15730.

\bibitem{tsipras2019robustness}
D.~Tsipras, S.~Santurkar, et~al., Robustness may be at odds with accuracy, in:
  International Conference on Learning Representations, 2019.

\bibitem{raghunathan2019adversarial}
A.~Raghunathan, S.~M. Xie, et~al., Adversarial training can hurt
  generalization, preprint arXiv:1906.06032 (2019).

\bibitem{croce2020reliable}
F.~Croce, M.~Hein, Reliable evaluation of adversarial robustness with an
  ensemble of diverse parameter-free attacks, in: International Conference on
  Machine Learning, PMLR, 2020, pp. 2206--2216.

\bibitem{cortes1995support}
C.~Cortes, V.~Vapnik, Support-vector networks, Machine learning (1995).

\bibitem{paszke2017automatic}
A.~P. et~al.,
  Pytorch: an imperative style, high-performance deep learning library,
  in: Advances in Neural Information Processing Systems, 2019.

\bibitem{isensee2021nnu}
F.~Isensee, P.~F. Jaeger, et~al., nnu-net: a self-configuring method for deep
  learning-based biomedical image segmentation, Nature Methods (2021).

\bibitem{daza2021towards}
L.~Daza, J.~C. P{\'e}rez, et~al., Towards robust general medical image
  segmentation, in: International Conference on Medical Image Computing and
  Computer-Assisted Intervention, Springer, 2021, pp. 3--13.

\bibitem{krizhevsky2009learning}
A.~Krizhevsky, G.~Hinton, Learning multiple layers of features from tiny
  images, in: Technical report, University of Toronto, Toronto, Ontario, 2009.

\bibitem{medmnistv2}
J.~Yang, R.~Shi, B.~Ni, Medmnist classification decathlon: A lightweight automl
  benchmark for medical image analysis (2021) 191--195.

\bibitem{soares2020sars}
E.~Soares, P.~Angelov, et~al., Sars-cov-2 ct-scan dataset: A large dataset of
  real patients ct scans for sars-cov-2 identification, preprint medRxiv
  (2020).

\bibitem{wu2018group}
Y.~Wu, K.~He, Group normalization, in: European Conference on Computer Vision,
  2018.

\bibitem{visser2020accurate}
M.~Visser, J.~Petr, et~al., Accurate mr image registration to anatomical
  reference space for diffuse glioma, Frontiers in Neuroscience (2020).

\bibitem{visser2019inter}
M.~Visser, D.~M{\"u}ller, et~al., Inter-rater agreement in glioma segmentations
  on longitudinal mri, NeuroImage: Clinical (2019).

\bibitem{cicchetti1994guidelines}
D.~V. Cicchetti, Guidelines, criteria, and rules of thumb for evaluating normed
  and standardized assessment instruments in psychology., Psychological
  Assessment (1994).

\bibitem{bartko1991measurement}
J.~J. Bartko, Measurement and reliability: statistical thinking considerations,
  Schizophr Bull (1991).

\bibitem{simpson2019large}
A.~L. Simpson, M.~Antonelli, et~al., A large annotated medical image dataset
  for the development and evaluation of segmentation algorithms, preprint
  arXiv:1902.09063 (2019).

\bibitem{jeong2022robust}
S.-w. Jeong, H.-h. Cho, et~al., Robust multimodal fusion network using
  adversarial learning for brain tumor grading, Computer Methods and Programs
  in Biomedicine 226 (2022) 107165.

\bibitem{qiu2022improved}
D.~Qiu, Y.~Cheng, et~al., Improved generative adversarial network for retinal
  image super-resolution, Computer Methods and Programs in Biomedicine 225
  (2022) 106995.

\bibitem{salvi2022dermocc}
M.~Salvi, F.~Branciforti, et~al., Dermocc-gan: A new approach for standardizing
  dermatological images using generative adversarial networks, Computer Methods
  and Programs in Biomedicine 225 (2022) 107040.

\bibitem{hazra2022enhancing}
D.~Hazra, Y.-C. Byun, et~al., Enhancing classification of cells procured from
  bone marrow aspirate smears using generative adversarial networks and
  sequential convolutional neural network, Computer Methods and Programs in
  Biomedicine 224 (2022) 107019.

\bibitem{pham2022generating}
Q.~T. Pham, S.~Ahn, et~al., Generating future fundus images for early
  age-related macular degeneration based on generative adversarial networks,
  Computer Methods and Programs in Biomedicine 216 (2022) 106648.

\end{thebibliography}
\end{singlespace}

\end{document}